\documentclass[12pt, draftclsnofoot, onecolumn]{IEEEtran}

\usepackage{amssymb}
\usepackage{amsmath}
\usepackage{cite}
\usepackage{url}
\usepackage{empheq}
\usepackage{xcolor}
\usepackage{graphicx}
\usepackage{subfigure}
\usepackage{enumitem}
\usepackage{fancyhdr}
\usepackage{mdwmath}	
\usepackage{mdwtab}
\usepackage{caption}
\usepackage{amsthm}
\usepackage{algorithm}
\usepackage{algorithmic}

\newtheorem{lemma}{Lemma}
\newtheorem{remark}{Remark}
\newtheorem{theorem}{Theorem}
\newtheorem{corollary}{Corollary}
\newtheorem{definition}{Definition}
\newtheorem{proposition}{Proposition}

\newcommand{\eqr}[1]{(\ref{#1})}
\newcommand{\fref}[1]{Fig.~\ref{#1}}

\newcommand*{\QEDA}{\null\nobreak\hfill\ensuremath{\blacksquare}}

\DeclareMathOperator*{\argmax}{argmax}

\begin{document}
\title{Convergence Acceleration in Wireless Federated Learning: A Stackelberg Game Approach}
\author{Kaidi~Wang,~\IEEEmembership{Member,~IEEE},
Yi~Ma,~\IEEEmembership{Senior Member,~IEEE},
Mahdi~Boloursaz~Mashhadi,~\IEEEmembership{Senior Member,~IEEE},
Chuan~Heng~Foh,~\IEEEmembership{Senior Member, IEEE},
Rahim~Tafazolli,~\IEEEmembership{Senior Member, IEEE},
and Zhi~Ding,~\IEEEmembership{Fellow, IEEE}
\thanks{K. Wang is with the Department of Electrical and Electronic Engineering, The University of Manchester, Manchester, M13 9PL, UK (email: kaidi.wang@ieee.org).}
\thanks{Y. Ma, M. Boloursaz Mashhadi, C. H. Foh, and R. Tafazolli are with 5GIC \& 6GIC, Institute for Communication Systems (ICS), University of Surrey, Guildford,  UK (email: y.ma@surrey.ac.uk; m.boloursazmashhadi@surrey.ac.uk; c.foh@surrey.ac.uk; r.tafazolli@surrey.ac.uk).}
\thanks{Z. Ding is with the Department of Electrical and Computer Engineering, University of California at Davis, Davis, CA 95616 USA (email: zding@ucdavis.edu).}}
\maketitle
\vspace{-14mm}\begin{abstract}
This paper studies issues that arise with respect to the joint optimization for convergence time in federated learning over wireless networks (FLOWN). We consider the criterion and protocol for selection of participating devices in FLOWN under the energy constraint and derive its impact on device selection. In order to improve the training efficiency, age-of-information (AoI) enables FLOWN to assess the freshness of gradient updates among participants. Aiming to speed up convergence, we jointly investigate global loss minimization and latency minimization in a Stackelberg game based framework. Specifically, we formulate global loss minimization as a leader-level problem for reducing the number of required rounds, and latency minimization as a follower-level problem to reduce time consumption of each round. By decoupling the follower-level problem into two sub-problems, including resource allocation and sub-channel assignment, we achieve an optimal strategy of the follower through monotonic optimization and matching theory. At the leader-level, we derive an upper bound of convergence rate and subsequently reformulate the global loss minimization problem and propose a new age-of-update (AoU) based device selection algorithm. Simulation results indicate the superior performance of the proposed AoU based device selection scheme in terms of the convergence rate, as well as efficient utilization of available sub-channels.
\end{abstract}
\section{Introduction}
The rapid development of mobile devices and applications has ushered us into the fifth-generation (5G) era. Much of the network services in 5G and beyond is expected to address explosive growth and need of machine learning (ML) and data science \cite{savazzi2021fl}. In conventional centralized ML, a central server is equipped at the access point (AP) to collect all raw data for model training. However, due to the limited wireless resources and potential privacy issues, centralized ML is impractical for some scenarios \cite{chen2021flm2}. In this context, federated learning (FL) is a framework for distributed ML algorithms to collaboratively train a central learning model while keeping the data locally \cite{qin2021fl}. Specifically, in FL, a global model is shared among multiple devices, and each device trains the received global model based on the local data and produces a local model \cite{xia2021fl}. Thereafter all local models are transmitted to the server via wireless communication networks to generate the updated global model \cite{chen2020flm}. Since raw data do not leave the device, and the model size is much smaller to share, such that there is less concern about data privacy and lower consumption of data network resources \cite{qiao2024massive}.
\subsection{Related Works}
Owing to its growing popularity, FL-related design and optimization in existing wireless communication architectures have attracted widespread attention, in which convergence time is regarded as an important performance metric. As indicated in \cite{konevcny2016federated}, the convergence time is jointly determined by the number of communication rounds and the time consumption per round. The former is closely related to convergence rate, while the latter is normally defined as latency.

In view of the relationship between global loss and convergence rate, some works focused on the global loss minimization problem in order to reduce the number of required communication rounds \cite{wang2019fl, chen2021fl1, hamdi2022fl, liu2022fl}. In \cite{wang2019fl}, a FL algorithm with multiple local training was designed. It considered the impact of local and global update rounds on the convergence bound, and developed an approximate solution of the global loss minimization problem. In \cite{chen2021fl1}, packet error rate was introduced to indicate whether the FL parameter transmission was successful or not. Specifically, the joint problem of user selection, resource block allocation, and power allocation was studied under delay and energy constraints \cite{chen2021fl1}. Incorporating FL in a massive multiple-input-multiple-output (MIMO) scenario with energy harvesting, the authors of \cite{hamdi2022fl} included the consideration of user scheduling and power allocation in a global loss minimization problem. Adopting a model pruning scheme, in \cite{liu2022fl}, the authors jointly optimized device selection, time slot allocation and pruning ratio in order to maximize the convergence rate with a latency constraint.

As another factor in determining the convergence time of FL, latency, including computation time and communication time, was extensively researched in previous works \cite{chen2021fl2, vu2020fl, chen2021flmatching, ji2021flmec}. The authors of \cite{chen2021fl2} designed a realistic wireless network for FL, where a limited number of users can be selected at each round for aggregation. By obtaining user selection and resource block allocation schemes, the convergence time of FL was minimized. Setting a local accuracy level at each device, the FL algorithm with multiple local update rounds was proposed in a cell-free massive MIMO scenario \cite{vu2020fl}, in which time consumption for downlink transmission, uplink transmission, and computation was considered in the formulated training time minimization problem. A multi-task FL framework was studied in a multi-access edge computing (MEC) scenario, where edge nodes were included to accomplish different learning tasks \cite{chen2021flmatching}. In order to minimize the latency of each communication round, the optimal matching between edge nodes and end devices was obtained. In \cite{ji2021flmec}, the authors proposed a hybrid learning scheme, where part of data can be offloaded from devices to the server, while the remaining data was utilized for local training.  It was demonstrated that the proposed scheme has the ability to reduce the total time consumption.
\subsection{Motivation and Contribution}
Although global loss minimization and latency minimization have been separately studied in existing works \cite{wang2019fl, chen2021fl1, hamdi2022fl, liu2022fl, chen2021fl2, vu2020fl, chen2021flmatching, ji2021flmec}, the interaction between these two objectives remains unclear. Specifically, devices that have a significant constructive impact on training convergence may have poor channel conditions, thereby increasing the latency of the corresponding aggregation round. On the other hand, focusing on minimizing latency may cause devices with high channel gains to be repeatedly selected, leading to an increase in the global loss \cite{nguyen2022contextual, imani2022fl}. Therefore, it is necessary to investigate this interaction and construct a dynamic trade-off. To this end, this work adopts Stackelberg game and presents a novel framework to jointly consider learning and communication in wireless FL systems, where the server and devices tend to minimize the global loss and latency, respectively. Unlike the conventional papers on global loss minimization that treat latency as a definite threshold \cite{wang2019fl, chen2021fl1, hamdi2022fl, liu2022fl}, latency in this work can be flexibly adjusted to ensure the convergence rate. Compared to \cite{shi2021fl}, an energy budget is included and its impact on device selection is analyzed, constructing a more practical and challenging scenario. On the other hand, inspired by the concept of age-of-information (AoI) \cite{yates2021aoi}, age-of-update (AoU) \cite{yang2020aoi} is defined in this work as a metric to evaluate the staleness of model updates. In this context, a novel device selection method is designed to estimate the contribution of devices in each communication round without analyzing the model/gradient or transmitting additional information to the server. Different from \cite{yang2020aoi} and \cite{kaidi2023fl2} which target overall AoU/AoI minimization, AoU in this work is regarded as a weight to prioritize selecting devices with larger AoU. The main contributions of this paper are summarized as follows:
\begin{itemize}[leftmargin=*]
\item A latency-sensitive FL scenario is considered, where multiple devices transmit parameters to the server over a limited number of sub-channels. In order to jointly minimize global loss and latency, a Stackelberg game based problem is formulated, where global loss minimization and latency minimization are considered as leader-level and follower-level problems, respectively. It is proved that the with the given sub-channels, some devices cannot transmit local models to the server due to the energy consumption constraint. Based on the analysis, the Stackelberg equilibrium of the formulated problems is established.
\item The follower-level problem is divided into two sub-problems, including resource allocation and sub-channel assignment. Due to non-convexity and monotonicity, a monotonic optimization based solution is proposed for the resource allocation problem. Moreover, a matching based algorithm is developed to address the sub-channel assignment problem with the incomplete preference list, where the properties of the proposed algorithm are analyzed.
\item In order to solve the leader-level problem, the upper bound of the convergence rate is derived, which indicates that the convergence rate can be improved by selecting devices with large data size. Therefore, the global loss minimization problem is reformulated as a weighted device selection problem. By ordering devices based on AoU and data size, a priority list is created, and an algorithm is designed to select devices by predicting sub-channel assignment and resource allocation.
\item The simulation results on Modified National Institute of Standards and Technology (MNIST), Canadian Institute for Advanced Research (CIFAR-10) and Stanford Sentiment Treebank Version 2 (SST-2) databases are presented. It is indicated that the designed AoU based device selection scheme can improve the convergence rate and achieve the lowest global loss. Moreover, the proposed solutions for resource allocation and sub-channel assignment can efficiently utilize available sub-channel and dynamically adjust energy utilization in order to reduce latency.
\end{itemize}
\subsection{Organization}
The remainder of this paper is organized as follows. In Section II and Section III, the system model and problem formulation are described, respectively. The solution of latency minimization problem is presented in Section IV, and the solution of global loss minimization problem is obtained in Section V. Section VI demonstrates the simulation results. The conclusions are summarized in Section VII.
\section{System Model}
Consider an FL scenario where $N$ wireless mobile devices collaboratively train a joint learning model. Each device is equipped with a single antenna and the FL process is orchestrated by a wireless server. In each communication round, the devices intend to train neural networks based on local data and then transmit parameters to the server for aggregation. Moreover, the limited communication resources are considered. Specifically, there are $K$ available sub-channels, $K\le N$, and each sub-channel is occupied by at most one device. Therefore, only a subset of devices can be selected for the global model aggregation in each communication round. The collections of all devices and sub-channels are denoted by $\mathcal{N}=\{1, 2, \dots, N\}$ and $\mathcal{K}=\{1, 2, \dots, K\}$, respectively.
\subsection{Computation Model}
In each communication round, after receiving the global model, the selected devices need to train their respective local learning models with the equipped central processing units (CPUs). Based on the dynamic voltage and frequency scaling (DVFS) technique, the CPU core can be operated at different frequency levels, and hence, the consumed time and energy change accordingly \cite{kaidi2021mec1}. For any device $n$ assigned to sub-channel $k$, the computational time consumption is given by
\begin{equation}\label{CPUrate}
T_{k,n}^{\mathrm{cp}}(\tau_{k,n}) = \frac{\mu\beta_n}{\tau_{k,n} C_n},
\end{equation}
where $\mu$ is a coefficient to denote the required CPU cycles for training one sample, $\beta_n$ is the number of dataset samples at device $n$, $\tau_{k,n}$ is a designed proportion of computational capacity, and $C_n$ is the CPU frequency of device $n$. Note that since the size of local data utilized in training does not change over the computing time, the test accuracy or loss reduction is not affected. The energy consumption for computation can be expressed as follows:
\begin{equation}
E_{k,n}^{\mathrm{cp}}(\tau_{k,n})  = \kappa_0\mu\beta_n(\tau_{k,n} C_n)^2,
\end{equation}
where $\kappa_0$ is the power consumption coefficient per CPU cycle.
\subsection{Communication Model}
After training, local models are transmitted from selected devices to the server. For any device $n$, the achievable data rate at sub-channel $k$ is given by
\begin{equation}
R_{k,n}(p_{k,n})=B\log_2(1+p_{k,n}|h_{k,n}|^2),
\end{equation}
where $B$ is the bandwidth, $p_{k,n}$ is the power allocation coefficient of device $n$, and $|h_{k,n}|^2$ is the normalized channel gain. Particularly, $|h_{k,n}|^2=P_t|g_{k,n}|^2\eta d_n^{-a}\sigma^{-2}$, where $P_t$ is the maximum transmit power in each sub-channel, $g_{k,n}\sim CN(0,1)$ the small-scale fading coefficient, $\eta$ is the frequency dependent factor, $d_n$ is the distance between device $n$ and the server, $a$ is the path loss exponent, and $\sigma^2$ is the variance of additive white Gaussian noise (AWGN). In the considered scenario, the small-scale fading coefficients vary with the communication rounds\footnote{Despite the location of devices is considered stationary in this paper, the proposed scheme can be extended to mobile scenarios, such as \cite{mahdi2023cl, zhang2024fl, zhang2024survey}.}. For simplicity, the index related to communication rounds is omitted from the notations. Based on the achievable data rate, the time consumption for communication can be presented by
\begin{equation}
T_{k,n}^{\mathrm{cm}}(p_{k,n})=\frac{D(\boldsymbol{w}_n^\mathrm{(t)})}{R_{k,n}(p_{k,n})},
\end{equation}
where $D(\boldsymbol{w}_n^\mathrm{(t)})$ is the size of the local model $\boldsymbol{w}_n^\mathrm{(t)}$ at device $n$ in round $t$. It is assumed that the data size of local models is the same for all devices and rounds, i.e., $D(\boldsymbol{w})=D(\boldsymbol{w}_n^\mathrm{(t)}), \forall n, t$. The energy consumption for communication is given by
\begin{equation}
E_{k,n}^{\mathrm{cm}}(p_{k,n}) = p_{k,n}P_tT_{k,n}^{\mathrm{cm}}(p_{k,n}).
\end{equation}
\subsection{AoU based Device Selection}
\begin{figure}[t]
\centering{\includegraphics[width=120mm]{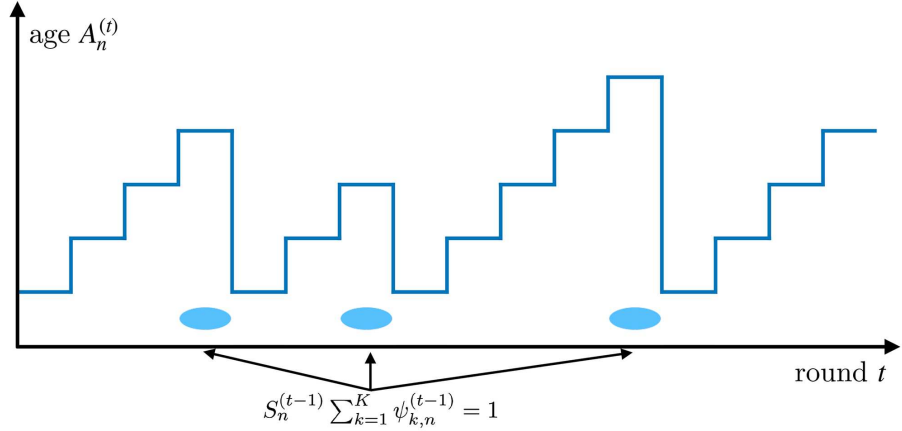}}
\caption{An illustration of device's information age.}
\label{ageinformation}
\end{figure}

In any communication round $t$, a subset of devices, denoted by $\mathcal{N}_t$, is selected,  i.e., $\mathcal{N}_t\subseteq\mathcal{N}$ and $|\mathcal{N}_t|\le K$. The status of any device in round $t$ can be represented by a binary variable $S_n^\mathrm{(t)}\in \{0, 1\}$, where $S_n^\mathrm{(t)}=1$ indicates device $n$ is selected to participate in the aggregation in round $t$; $S_n^\mathrm{(t)}=0$ otherwise. The set of all device selection indicators is denoted by $\mathbf{S}^\mathrm{(t)}$. Therefore, in any round, the set of selected devices $\mathcal{N}_t$ and the set of device selection indicators $\mathbf{S}^\mathrm{(t)}$ can be mutually inferred. 

In order to improve the performance of FL, device selection should be determined by data quality. It is revealed in \cite{dai2018toward, yates2021aoi} that staleness of gradient update may negatively impact learning outcomes. Therefore, the server tends to select the devices with fresher gradient update for aggregation. In particular, if a device has been skipped for several rounds, its gradient update is relatively informative, and the probability of the server selecting the device should increase. Conversely, if a device was selected in the previous round, the probability for its reselection should decrease. To this end, the concept of AoI is adopted to define AoU \cite{talak2020aoi, yang2020aoi}. In round $t$, any device $n$'s AoU, denoted by $A_n^\mathrm{(t)}$, is presented as follows:
\begin{equation}
A_n^\mathrm{(t)}=\left\{\begin{array}{ll}
A_n^\mathrm{(t-1)}+1, &\!\!\!\!\!\quad\text{if}\quad S_n^\mathrm{(t-1)}\sum_{k=1}^{K}\psi_{k,n}^\mathrm{(t-1)}=0,\\
1, &\!\!\!\!\!\quad\text{if}\quad S_n^\mathrm{(t-1)}\sum_{k=1}^{K}\psi_{k,n}^\mathrm{(t-1)}=1,
\end{array}\right.
\end{equation}
where $\psi_{k,n}^\mathrm{(t)}\in\{0, 1\}$ is the sub-channel assignment indicator. Particularly, $\psi_{k,n}^\mathrm{(t)}=1$ indicates that device $n$ is assigned to sub-channel $k$ in round $t$; $\psi_{k,n}^\mathrm{(t)}=0$ otherwise. According to the above definition, AoU represents the number of communication rounds since the last transmission, which is jointly decided by device selection and sub-channel assignment. As shown in \fref{ageinformation}, if device $n$ was not selected in round $t-1$, i.e., $S_n^\mathrm{(t-1)}=0$, or it is not assigned to any sub-channel, i.e., $\sum_{k=1}^{K}\psi_{k,n}^\mathrm{(t-1)}=0$, its AoU is incremented by $1$; otherwise, its AoU is reset to $1$. With this consideration, a large AoU implies more informative update, and therefore device selection should partially depend on AoU. In any round $t$, $\alpha_n^\mathrm{(t)}$ defined below is used as a weight\footnote{In this work, the term ``weight'' refers to the weighting coefficient, and the neural network weights are represented by the term ``model''.} to prioritize selecting devices with larger AoU for aggregation:
\begin{equation}
\alpha_n^\mathrm{(t)} = \frac{A_n^\mathrm{(t)}}{\sum_{i=1}^N A_i^\mathrm{(t)}}.
\end{equation}
\subsection{Sub-Channel Assignment}
The time consumption of each communication round, i.e., latency, also plays an important role in FL \cite{konevcny2016federated, ren2020fl, chen2021fl2}. By including computation and communication phases, the time consumption of any device $n$ assigned to sub-channel $k$ can be expressed as follows:
\begin{equation}
T_{k,n}(\tau_{k,n}, p_{k,n})=T_{k,n}^{\mathrm{cp}}(\tau_{k,n})+T_{k,n}^{\mathrm{cm}}(p_{k,n}),
\end{equation}
where the time consumption for global model transmission from the server to devices is ignored as in \cite{chen2021fl3, yang2021fl}. With the given set of selected devices $\mathcal{N}_t$, the latency in round $t$ can be presented as follows:
\begin{equation}
T^\mathrm{(t)}=\max_{n\in\mathcal{N}_t}\left\{\sum_{k=1}^K\psi_{k,n}^\mathrm{(t)}T_{k,n}(\tau_{k,n}, p_{k,n})\right\}.
\end{equation}
It indicates that the latency is affected by sub-channel assignment. Specifically, if any device is assigned to a sub-channel with an enhanced channel gain, the time consumption of this device decreases, and hence, the latency of this round may be reduced. Accordingly, the energy consumption of device $n$ assigned to sub-channel $k$ is given by
\begin{equation}
E_{k,n}(\tau_{k,n}, p_{k,n}) = E_{k,n}^{\mathrm{cp}}(\tau_{k,n})+E_{k,n}^{\mathrm{cm}}(p_{k,n}).
\end{equation}
\section{Problem Formulation}
This work focuses on minimizing the convergence time of FL, which is defined as the sum of latency across all communication rounds \cite{konevcny2016federated}. Therefore, the convergence time is determined by i) the number of communication rounds, and ii) the time consumption of each round. That is, the server can select devices with more data and larger AoU to reduce the number of required rounds, or devices with better channel conditions to reduce the latency in each round. However, devices with more data and larger AoU are often not preferred due to larger time and energy consumption, while devices with good channel conditions can be frequently selected, which usually causes smaller AoU. Therefore, there exists a trade-off between these two options. In order to comprehensively minimize convergence time, the number of required communication rounds and latency should be jointly minimized. This situation can be described by the Stackelberg game, where global loss minimization and latency minimization are considered as leader-level and follower-level problems, respectively.
\subsubsection{Leader-Level Problem (Computation)}
In the considered FL algorithm, the local loss at device $n$ is given by
\begin{equation}
f_n(\boldsymbol{w}^\mathrm{(t)}) = \frac{1}{\beta_n}\sum_{i=1}^{\beta_n} \ell(\boldsymbol{w}^\mathrm{(t)}; \boldsymbol{x}_{n,i}, y_{n,i}),
\end{equation}
where $\boldsymbol{w}^\mathrm{(t)}$ is the global model in round $t$, $\ell(\boldsymbol{w}^\mathrm{(t)}; \boldsymbol{x}_{n,i}, y_{n,i})$ is a loss function, $\boldsymbol{x}_{n,i}$ is the $i$-th input data of device $n$, and $y_{n,i}$ is the corresponding label. In round $t$, the global loss can be presented below
\begin{equation}
F(\boldsymbol{w}^\mathrm{(t)})=\frac{\sum_{n=1}^N\sum_{i=1}^{\beta_n} \ell(\boldsymbol{w}^\mathrm{(t)}; \boldsymbol{x}_{n,i}, y_{n,i})}{\sum_{n=1}^N\beta_n}.
\end{equation}
By including device selection and sub-channel assignment, an AoU based global loss minimization problem is formulated as follows:
\begin{align}
\min_{\mathbf{S}^\mathrm{(t)}} \quad & \frac{\sum_{n=1}^N\!\alpha_n^\mathrm{(t)}S_n^\mathrm{(t)}\!\sum_{k=1}^K\!\psi_{k,n}^\mathrm{(t)}\!\sum_{i=1}^{\beta_n}\ell(\boldsymbol{w}^\mathrm{(t)}; \boldsymbol{x}_{n,i}, y_{n,i})}{\sum_{n=1}^N\!\alpha_n^\mathrm{(t)}S_n^\mathrm{(t)}\!\sum_{k=1}^K\!\psi_{k,n}^\mathrm{(t)}\beta_n}, \!\label{lproblem}\\\nonumber
\textrm{s.t.} \quad & S_n^\mathrm{(t)}\in\{0, 1\}, \forall n \in\mathcal{N}, \tag{\ref{lproblem}a}\\
& \sum\nolimits_{n=1}^N S_n^\mathrm{(t)} \le K.\tag{\ref{lproblem}b}
\end{align}
Constraint  (\ref{lproblem}b) indicates that the number of selected devices in each round is not greater than the number of available sub-channels. Note that only device selection is considered in the computation phase, and the sub-channel assignment indicator is decided in the communication phase, even though it has an impact on the global loss.
\subsubsection{Follower-Level Problem (Communication)}
Based on the $\mathbf{S}^\mathrm{(t)}$ given by the leader-level problem, the set of selected devices $\mathcal{N}_t$ can be obtained. Then, sub-channel assignment and resource allocation can be implemented according to the set $\mathcal{N}_t$. In any round $t$, the latency minimization problem can be presented as follows:
\begin{align}
\min_{\boldsymbol{\psi}^\mathrm{(t)}, \boldsymbol{\tau}, \boldsymbol{p}} \quad & \max_{n\in\mathcal{N}_t}\left\{\sum_{k=1}^K\psi_{k,n}^\mathrm{(t)}T_{k,n}(\tau_{k,n}, p_{k,n})\right\},\label{fproblem}\\\nonumber
\textrm{s.t.} \quad & E_{k,n}(\tau_{k,n}, p_{k,n})\le E_{n}^{\mathrm{max}},\tag{\ref{fproblem}a}\\
& \tau_{k,n}\in[0, 1], p_{k,n}\in[0, 1], \tag{\ref{fproblem}b}\\
& \psi_{k,n}^\mathrm{(t)}\in\{0, 1\}, \tag{\ref{fproblem}c}\\
& \sum\nolimits_{n\in \mathcal{N}_t}\psi_{k,n}^\mathrm{(t)} =1, \forall k\in\mathcal{K}, \tag{\ref{fproblem}d}\\
& \sum\nolimits_{k=1}^K \psi_{k,n}^\mathrm{(t)}=1, \forall n\in\mathcal{N}_t,\tag{\ref{fproblem}e}
\end{align}
where $\boldsymbol{\psi}^\mathrm{(t)}$, $\boldsymbol{\tau}$, and $\boldsymbol{p}$ are the sets of all sub-channel assignment indicators, computational resource allocation coefficients, and power allocation coefficients, respectively. In constraint (\ref{fproblem}a), the maximum energy consumption $E_{n}^{\text{max}}$ is included. In constraints (\ref{fproblem}b) and  (\ref{fproblem}c), the value ranges of all optimization variables are defined. Constraints (\ref{fproblem}d) and (\ref{fproblem}e) indicate that each sub-channel can be occupied by one device, and each device can be assigned to one sub-channel, respectively.
\subsubsection{Stackelberg Equilibrium}
With any given solution of the leader-level problem, the formulated follower-level problem can be infeasible, as shown in follows:
\begin{proposition}\label{infeasible}
With any device selection $S_n^\mathrm{(t)}=1$, problem \eqr{fproblem} is infeasible if the following condition holds:
\emph{\begin{equation}
\ln(2)P_tD(\boldsymbol{w}_n^\mathrm{(t)})\ge E_{n}^{\mathrm{max}}B|h_{k,n}|^2, \forall k\in\mathcal{K}.
\end{equation}}
\begin{IEEEproof}
Refer to Appendix~A.	
\end{IEEEproof}
\end{proposition}
Proposition \ref{infeasible} indicates that the energy consumption constraint can affect device participation, thereby reducing learning performance \cite{kaidi2024fl1}. The following remark can be obtained.
\begin{remark}
In wireless FL scenarios, energy consumption constraints can restrict the transmission of local models individually, resulting in a decrease in the global loss.
\end{remark}
The above proposition and remark show that there exists an interaction between these two problems. That is, the selected devices may not be able to transmit local models to the server due to the poor channel conditions\footnote{Note that transmit power $P_t$ does not affect the feasibility as the normalized channel condition $|h_{k,n}|^2$ also contains the transmit power.}. This interaction is consistent with the Stackelberg competition model, in which the leader and the follower tend to maximize their own utilities \cite{han2012game}. For solving the formulated Stackelberg game based problem, Stackelberg equilibrium \cite{basar1999dynamic} is introduced as follows:
\begin{definition}
In the formulated Stackelberg game based problem, by respectively defining $G_\mathrm{L}$ and $G_\mathrm{F}$ as the objective functions of leader-level and follower-level problems, solution $(\mathbf{S}^{\mathrm{(t)}*},\boldsymbol{\psi}^{\mathrm{(t)}*}, \boldsymbol{\tau}^*, \boldsymbol{p}^*)$ is the Stackelberg equilibrium if the following conditions hold:
\begin{equation}
\begin{array}{ll}
G_\mathrm{L}(\mathbf{S}^{\mathrm{(t)}*},\boldsymbol{\psi}^{\mathrm{(t)}*}, \boldsymbol{\tau}^*, \boldsymbol{p}^*) &\!\!\!\!\le G_\mathrm{L}(\mathbf{S}^\mathrm{(t)},\boldsymbol{\psi}^{\mathrm{(t)}*}, \boldsymbol{\tau}^*, \boldsymbol{p}^*),\\
G_\mathrm{F}(\mathbf{S}^{\mathrm{(t)}*}, \boldsymbol{\psi}^{\mathrm{(t)}*}, \boldsymbol{\tau}^*, \boldsymbol{p}^*) &\!\!\!\!\le G_\mathrm{F}(\mathbf{S}^{\mathrm{(t)}*},\boldsymbol{\psi}^\mathrm{(t)}, \boldsymbol{\tau}, \boldsymbol{p}).
\end{array}
\end{equation}
\end{definition}
In order to achieve the Stackelberg equilibrium, the leader should predict the possible solution of the follower-level problem, i.e., the feasibility of the selected devices, and then propose a strategy which can minimize the global loss after transmission. As a result, the solution of the follower-level problem should be obtained before the leader-level problem. Note that both problems are solved at the server, and the solutions of each communication round can be transmitted to all devices together with the global model. Since the server has powerful computing capabilities, the impact of this process on latency can be ignored.
\section{Solution of Follower-Level Problem}
The formulated follower-level problem in \eqr{fproblem} is a non-convex problem with binary constraints. In this section, the follower-level problem is decoupled into two sub-problems and solved iteratively. With the fixed sub-channel assignment, the sub-problem related to resource allocation is given by
\begin{align}
\boldsymbol{\Gamma}=\min_{\boldsymbol{\tau}, \boldsymbol{p}} \quad & \left\{T_{k,n}(\tau_{k,n}, p_{k,n})| \forall n\in\mathcal{N}_t, \forall k\in\mathcal{K}\right\}, \label{subproblem1}\\\nonumber
\textrm{s.t.} \quad & \text{(\ref{fproblem}a), (\ref{fproblem}b)},
\end{align}
where $\boldsymbol{\Gamma}$ is a $K\times |\mathcal{N}_t|$ matrix containing the minimum time consumptions for all possible device and sub-channel combinations. With the given matrix $\boldsymbol{\Gamma}$, the sub-problem related to sub-channel assignment can be presented as follows:
\begin{align}
\min_{\boldsymbol{\psi}^\mathrm{(t)}} \quad & \max_{n\in\mathcal{N}_t}\left\{\sum_{k=1}^K\psi_{k,n}^\mathrm{(t)}\Gamma_{k,n}\right\},\label{subproblem2}\\\nonumber
\textrm{s.t.} \quad & \text{(\ref{fproblem}c), (\ref{fproblem}d), (\ref{fproblem}e)},
\end{align}
where $\Gamma_{k,n}$ is one element of matrix $\boldsymbol{\Gamma}$. Note that there exists an infeasible combination if the condition in Proposition \ref{infeasible} holds. In this case, this combination will be marked as infeasible and avoided in problem \eqr{subproblem2}.
\subsection{Joint Optimization of Computational Resource Allocation and Power Allocation}
In problem \eqr{subproblem1}, the time consumption of any combination only depends on its corresponding resource allocation. Therefore, this problem can be divided into multiple sub-problems to obtain all elements of matrix $\boldsymbol{\Gamma}$. The feasibility of combinations is verified by utilizing Proposition \ref{infeasible}. For any feasible combination $\Gamma_{k,n}$, the sub-problem is given by
\begin{align}
\min_{\tau_{k,n}, p_{k,n}} \!\quad\! & T_{k,n}^{\mathrm{cp}}(\tau_{k,n})+T_{k,n}^{\mathrm{cm}}(p_{k,n})\label{subproblem11}\\
\textrm{s.t.} \quad & E_{k,n}^{\mathrm{cp}}(\tau_{k,n})+E_{k,n}^{\mathrm{cm}}(p_{k,n}) \le E_{n}^\mathrm{max}, \tag{\ref{subproblem11}a}\\
& \tau_{k,n} \in[0, 1], p_{k,n} \in[0, 1]. \tag{\ref{subproblem11}b}
\end{align}
Since problem \eqr{subproblem11} is non-convex, traditional optimization methods, such as convex optimization, cannot be directly employed. In this case, monotonic optimization is introduced. In order to utilize monotonic optimization, the monotonicity of problem \eqr{subproblem11} should be analyzed. According to \cite{tuy2000monotonic}, the following proposition can be obtained.
\begin{proposition}\label{monotonic}
For any device $n$ assigned to sub-channel $k$, with computational resource allocation coefficient $\tau_{k,n}$ and power allocation coefficient $p_{k,n}$, the time consumption $T_{k,n}(\tau_{k,n}, p_{k,n})$ is a decreasing function, while the energy consumption $E_{k,n}(\tau_{k,n}, p_{k,n})$ is an increasing function.
\begin{IEEEproof}
Refer to Appendix~B.	
\end{IEEEproof}
\end{proposition}
Proposition \ref{monotonic} indicates the following remark.
\begin{remark}
In the considered FL framework, minimizing the latency leads to the maximized energy consumption.
\end{remark}
According to Proposition \ref{monotonic}, it is indicated that the objective function and all constraints in problem \eqr{subproblem11} are monotonic. Next, this problem is transformed to the canonical formulation, as shown in follows:
\begin{align}
\max_{\mathbf{z}_{k,n}} \quad & f(\mathbf{z}_{k,n}) \label{subproblem13}\\
\textrm{s.t.} \quad & \mathbf{z}_{k,n}\in \mathcal{G}, \tag{\ref{subproblem13}a}
\end{align}
where $\mathbf{z}_{k,n}=\{\tau_{k,n}, p_{k,n}\}$, $\mathcal{G}=\{\mathbf{z}_{k,n}\in[\mathbf{0}, \mathbf{1}], g(\mathbf{z}_{k,n})\le 0\}$,
\begin{equation}\label{functionz}
f(\mathbf{z}_{k,n})=-\frac{\mu \beta_n}{\tau_{k,n} C_n}-\frac{D(\boldsymbol{w}_n^\mathrm{(t)})}{B\log_2(1+p_{k,n}|h_{k,n}|^2)},
\end{equation}
and
\begin{equation}
g(\mathbf{z}_{k,n})\!=\!\kappa_0\mu \beta_n(\tau_{k,n} C_n)^2\!+\!\frac{p_{k,n}P_tD(\boldsymbol{w}_n^\mathrm{(t)})}{B\log_2(1\!+\!p_{k,n}|h_{k,n}|^2)}\!-\!E_{n}^\mathrm{max}.
\end{equation}
In problem \eqr{subproblem13}, the objective function $f(\mathbf{z}_{k,n})$ is monotonically increasing with $\mathbf{z}_{k,n}$, and then the optimal solution is located on the boundary of the feasible set $\mathcal{G}$. However, due to the non-convexity, the expression of the boundary cannot be directly derived. In this case, a polyblock outer approximation algorithm is proposed to approach the feasible set $\mathcal{G}$ by constructing polyblock $\mathcal{P}$ \cite{zhang2013monotonic}, as shown in \textbf{Algorithm \ref{alg1}}.

\begin{algorithm}[t]
\caption{Polyblock Outer Approximation Algorithm}
\label{alg1}
\begin{algorithmic}[1]
\STATE Initialize vertex set $\mathcal{V}^{(1)}=\{\mathbf{v}^{(1)}\}$, where $\mathbf{v}^{(1)}=\{1, 1\}$.
\STATE Initialize polyblock $\mathcal{P}^{(1)}$ with vertex set $\mathcal{V}^{(1)}$.
\STATE Set $\epsilon$ and $\theta=1$.
\IF{$|f(\boldsymbol{\phi}(\mathbf{v}^{(\theta)}))-f(\boldsymbol{\phi}(\mathbf{v}^{(\theta-1)}))|>\epsilon$}
\STATE Obtain $\boldsymbol{\phi}(\mathbf{v}^{(\theta)})$ from Eq. \eqr{projection}.
\STATE Calculate vertices $\tilde{\mathbf{v}}_1^{(\theta)}$ and $\tilde{\mathbf{v}}_2^{(\theta)}$ as follows:
\vspace{-1mm}\begin{equation}\vspace{-1mm}\nonumber
\tilde{\mathbf{v}}_i^{(\theta)}=\mathbf{v}^{(\theta)}-(v_i^{(\theta)}-\phi_i(\mathbf{v}^{(\theta)}))\mathbf{e}_i, \forall i\in\{1,2\}.
\end{equation}
\STATE Update vertex set $\mathcal{V}^{(\theta+1)}$ as follows:
\vspace{-1mm}\begin{equation}\vspace{-1mm}\nonumber
\mathcal{V}^{(\theta+1)}=\{\mathcal{V}^{(\theta)}\backslash\mathbf{v}^{(\theta)}\}\cup\{\mathbf{\tilde{v}}_1^{(\theta)}, \mathbf{\tilde{v}}_2^{(\theta)}\}.
\end{equation}
\STATE Construct polyblock $\mathcal{P}^{(\theta+1)}$ with vertex set $\mathcal{V}^{(\theta+1)}$.
\STATE Find vertex $\mathbf{v}^{(\theta+1)}$ from $\mathcal{V}^{(\theta+1)}$, where
\vspace{-1mm}\begin{equation}\vspace{-1mm}\nonumber
\mathbf{v}^{(\theta+1)}=\argmax\{f(\boldsymbol{\phi}(\mathbf{v}))|\mathbf{v}\in\mathcal{V}^{(\theta+1)}\}.
\end{equation}
\STATE Set $\theta=\theta+1$.
\ENDIF
\STATE Set $\mathbf{z}_{k,n}^*=\boldsymbol{\phi}(\mathbf{v}^{(\theta)})$.
\end{algorithmic}
\end{algorithm}

\begin{figure}[t]
\centering{
\subfigure[]{\centering{\includegraphics[width=35mm]{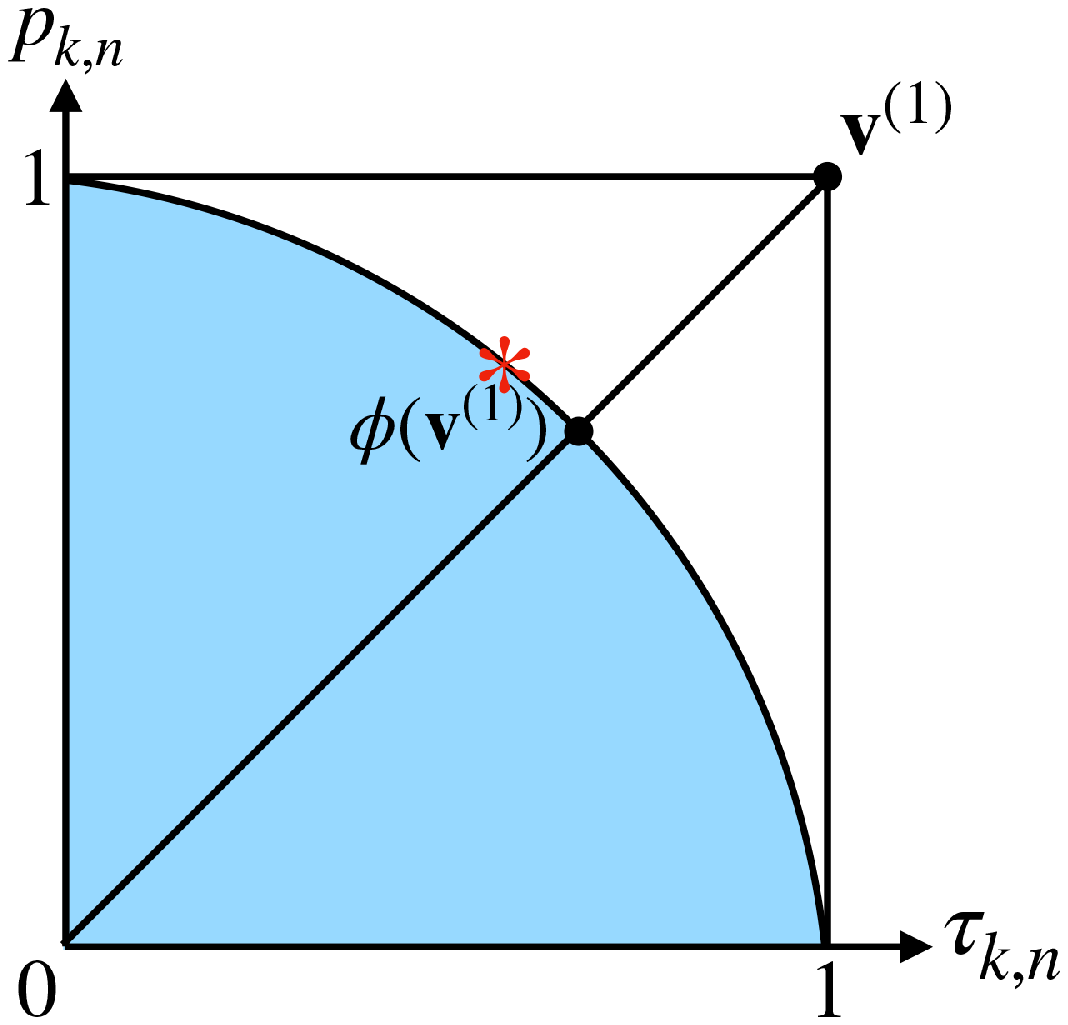}}}
\subfigure[]{\centering{\includegraphics[width=35mm]{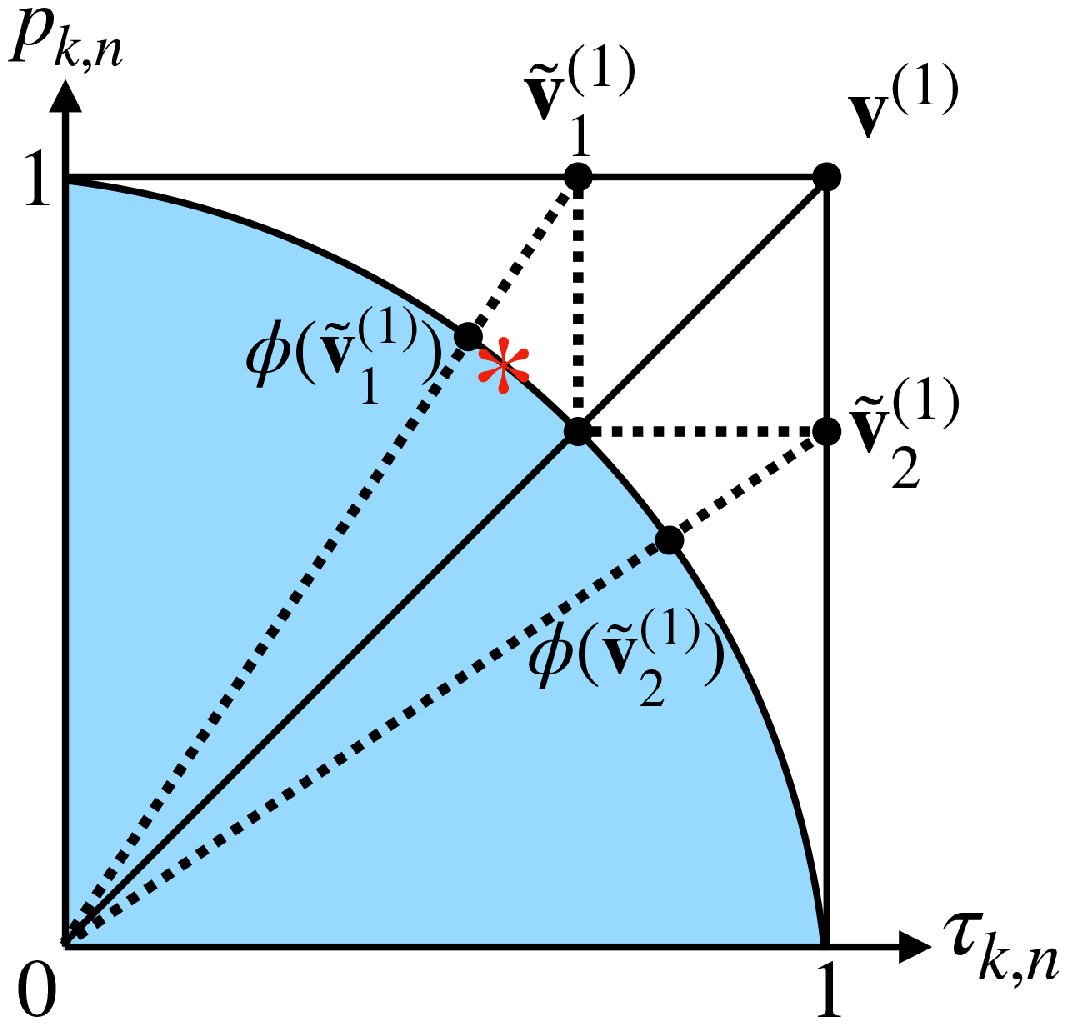}}}
\subfigure[]{\centering{\includegraphics[width=35mm]{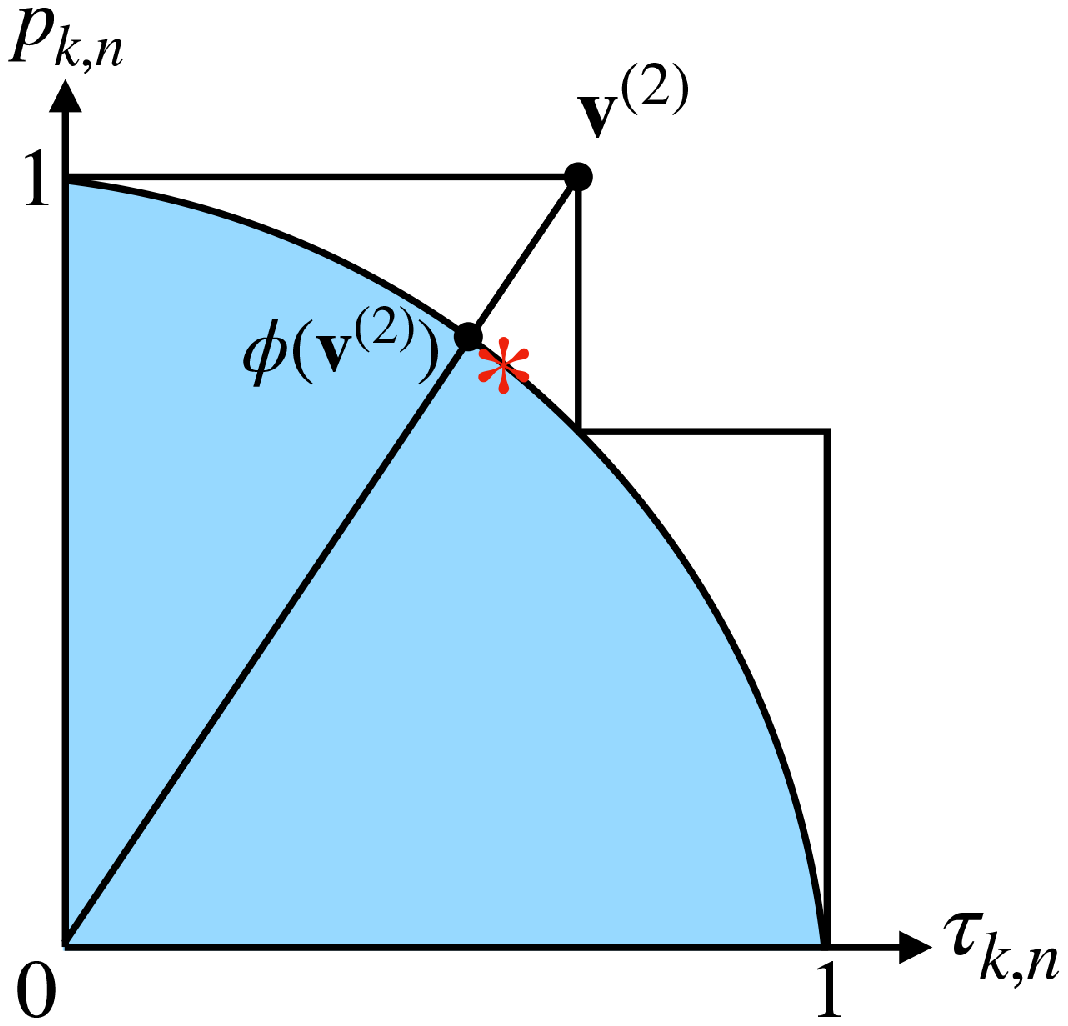}}}
\subfigure[]{\centering{\includegraphics[width=35mm]{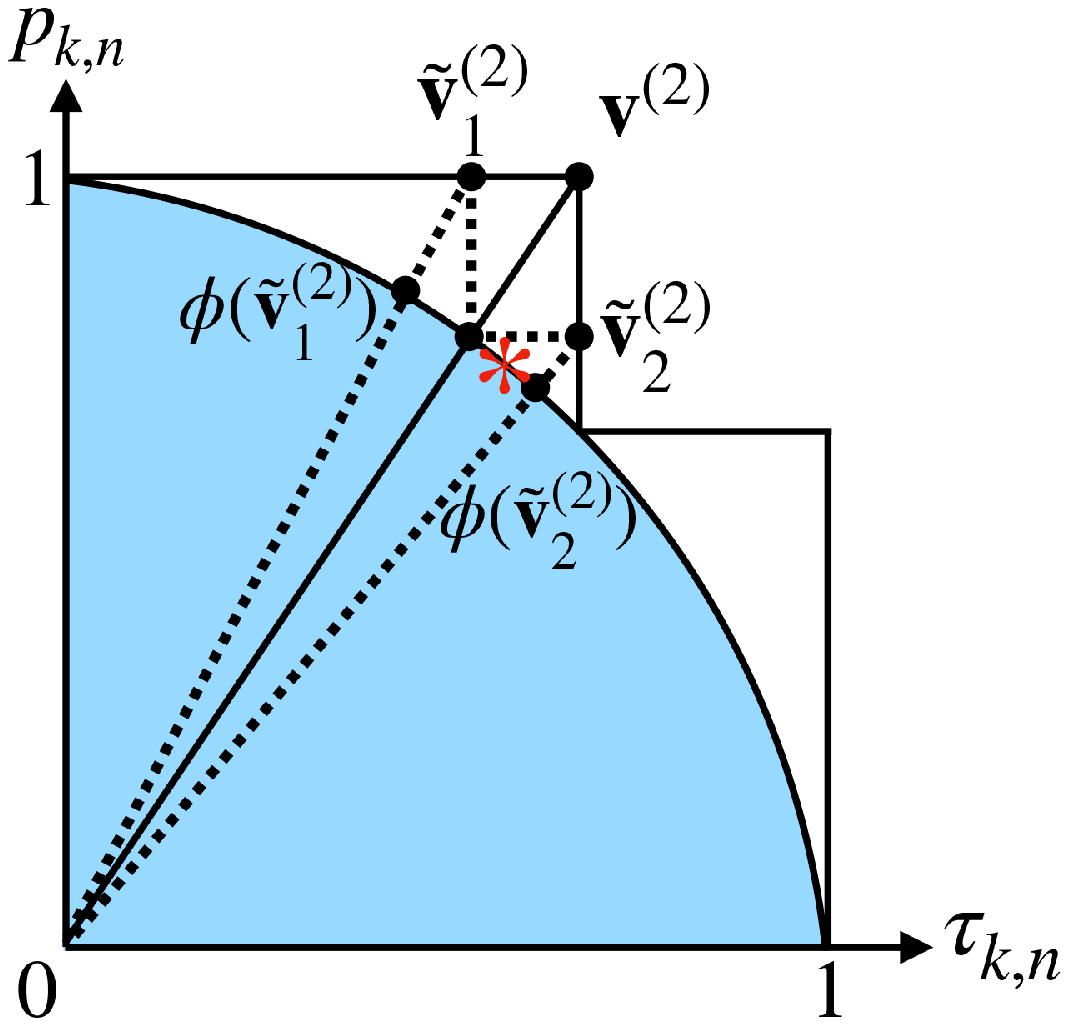}}}}
\caption{An illustration of Algorithm 1. The blue area is the feasible set $\mathcal{G}$, and the red star is the optimal point.}
\label{mo}
\vspace{-6mm}
\end{figure}

As shown in \fref{mo}(a), by setting the first vertex $\mathbf{v}^{(1)}\in\mathcal{V}^{(1)}$, the initial polyblock $\mathcal{P}^{(1)}$ can be constructed as the box $[\mathbf{0},\mathbf{1}]$, which contains the feasible set\footnote{Note that the initial vertex, i.e., $\tau_{k,n}=1$ and $p_{k,n}=1$, may be infeasible for problem \eqr{subproblem13}. However, the optimal solution obtained from Algorithm \ref{alg1} is the projection of the vertex, which is always included in the feasible set $\mathcal{G}$.}. The projection of $\mathbf{v}^{(1)}$ on the upper boundary of feasible set $\mathcal{G}$ is calculated, denoted by $\boldsymbol{\phi}(\mathbf{v}^{(1)})$. Based on point $\boldsymbol{\phi}(\mathbf{v}^{(1)})$, two new vertices $\mathbf{\tilde{v}}_1^{(1)}$ and $\mathbf{\tilde{v}}_2^{(1)}$ can be obtained to replace $\mathbf{v}^{(1)}$, as shown in \fref{mo}(b). The new vertices are calculated as follows:
\begin{equation}
\tilde{\mathbf{v}}_i^{(1)}=\mathbf{v}^{(1)}-(v_i^{(1)}-\phi_i(\mathbf{v}^{(1)}))\mathbf{e}_i, \forall i\in\{1,2\},
\end{equation}
where $v_i^{(1)}$ is the $i$-th element of $\mathbf{v}^{(1)}$, $\phi_i(\mathbf{v}^{(1)})$ is the $i$-th element of $\boldsymbol{\phi}(\mathbf{v}^{(1)})$, and $\mathbf{e}_i$ is the $i$-th unit vector. As shown in \fref{mo}(c), a new polyblock $\mathcal{P}^{(2)}$ is obtained, where $\mathcal{G}\subset\mathcal{P}^{(2)}\subset\mathcal{P}^{(1)}$. The vertex set of $\mathcal{P}^{(2)}$ is updated as follows:
\begin{equation}
\mathcal{V}^{(2)}=\{\mathcal{V}^{(1)}\backslash\mathbf{v}^{(1)}\}\cup\{\mathbf{\tilde{v}}_1^{(1)}, \mathbf{\tilde{v}}_2^{(1)}\}.
\end{equation}
After that, the optimal vertex is selected from vertex set $\mathcal{V}^{(2)}$ and denoted by $\mathbf{v}^{(2)}$, which satisfies the following condition:
\begin{equation}
\mathbf{v}^{(2)}=\argmax\{f(\boldsymbol{\phi}(\mathbf{v}))|\mathbf{v}\in\mathcal{V}^{(2)}\}.
\end{equation}
The projection of the optimal vertex can achieve the maximum value of \eqr{functionz}. For example, \fref{mo}(c) shows the case that $\mathbf{\tilde{v}}_1^{(1)}$ is selected as the optimal vertex $\mathbf{v}^{(2)}$. This process is repeated, and a smaller polyblock $\mathcal{P}^{(3)}\subset\mathcal{P}^{(2)}$ is constructed based on vertex $\mathbf{v}^{(2)}$, as shown in \fref{mo}(d). This algorithm is completed if the following condition is satisfied:
\begin{equation}
|f(\boldsymbol{\phi}(\mathbf{v}^{(\theta)}))-f(\boldsymbol{\phi}(\mathbf{v}^{(\theta-1)}))| \le \epsilon,
\end{equation}
where $\epsilon$ is the error tolerance. The output $\mathbf{z}_{k,n}^*=\boldsymbol{\phi}(\mathbf{v}^{(\theta)})$ is the optimal solution of problem \eqr{subproblem13}. Moreover, the projection of any vertex $\mathbf{v}^{(\theta)}$ satisfies the following condition:
\begin{equation}
\boldsymbol{\phi}(\mathbf{v}^{(\theta)})=\zeta\mathbf{v}^{(\theta)},
\end{equation}
where $\zeta\in(0,1)$ is a ratio coefficient obtained as follows:
\begin{equation}
\zeta =\max\{\tilde{\zeta}\in(0,1)|\tilde{\zeta}\mathbf{v}^{(\theta)}\in\mathcal{G}\}=\max\{\tilde{\zeta}\in(0,1)|g(\tilde{\zeta}\mathbf{v}^{(\theta)})\le 0, \tilde{\zeta}\mathbf{v}^{(\theta)}\in[\mathbf{0}, \mathbf{1}]\}.
\end{equation}
Since $g(\tilde{\zeta}\mathbf{v}^{(\theta)})$ is a monotonic increasing function, $\zeta$ satisfies $g(\zeta\mathbf{v}^{(\theta)})=0$, which can be written as follows:
\begin{equation}\label{projection}
\kappa_0\mu \beta_n(\zeta v_1^{(\theta)} C_n)^2\!+\!\frac{\zeta v_2^{(\theta)}P_tD(\boldsymbol{w}_n^\mathrm{(t)})}{B\log_2(1\!+\!\zeta v_2^{(\theta)}|h_{k,n}|^2)}=E_{n}^\mathrm{max}.
\end{equation}
The above nonlinear equation can be solved by multiple numerical solvers, such as fsolve in MATLAB or Python. It is worth pointing out that during \textbf{Algorithm \ref{alg1}}, with any given vertex $\mathbf{v}^{(\theta)}=\{v_1^{(\theta)}, v_2^{(\theta)}\}$, where $\theta \ge 2$, condition $\zeta\in(0,1)$ is always satisfied. When $\theta = 1$, $\zeta=1$ may be obtained with initial vertex $\mathbf{v}^{(1)}$, which means that the feasible set $\mathcal{G}$ includes the first vertex. In other words, the energy consumption constraint $g(\mathbf{z}_{k,n})<0$ can be satisfied with any $\tau_{k,n}$ and $p_{k,n}$ in $[0, 1]$. In this case, the optimal solution is obtained by $\mathbf{z}_{k,n}^* = \mathbf{v}^{(1)}$.

According to \cite{hua2020mo}, the complexity of \textbf{Algorithm \ref{alg1}} is dominated by step 9 and the number of iterations. Specifically, step 9 selects an optimal vertex from the vertex candidate set, which contains $\theta+1$ vertices at the $\theta$-th iteration. Since the complexity in solving \eqr{subproblem13} to obtain each vertex is $\mathcal{O}(2)$, the complexity of step 9 is $\mathcal{O}(2(\theta+1))$. With the given number of iterations $T_c$, the total complexity of \textbf{Algorithm \ref{alg1}} can be expressed as $\mathcal{O}(2T_c(T_c+1))$.
\subsection{Matching based Sub-Channel Assignment}
Based on \textbf{Algorithm 1}, the minimum time consumption of all selected devices in all sub-channels is obtained in matrix $\boldsymbol{\Gamma}$. In this subsection, a matching based algorithm is proposed to solve the binary integer programming problem in \eqr{subproblem2}, where matrix $\boldsymbol{\Gamma}$ is utilized to construct the preference list. In problem \eqr{subproblem2}, the set of selected devices, i.e., $\mathcal{N}_t$, is provided by the leader-level problem. As revealed in Section \ref{leadersolution}, to minimize global loss, the number of selected devices is maximized in each round, i.e., $|\mathcal{N}_t|=K$. Therefore, $\mathcal{N}_t$ and $\mathcal{K}$ are two disjoint sets with the same size, and this scenario can be considered as a one-to-one matching $\Psi$ from $\mathcal{N}_t$ to $\mathcal{K}$. Furthermore, since some combinations in matrix $\boldsymbol{\Gamma}$ are marked as infeasible, the player in this matching may have an incomplete preference list \cite{gusfield1989stable, iwama2002stable}. In this case, the elements in $\boldsymbol{\Gamma}$ cannot be directly employed, and the utility of device $n$ assigned to sub-channel $k$ in matching $\Psi$ is defined below
\begin{equation}\label{utility}
U_n(\Psi)=\left\{
\begin{array}{ll}
U_{\mathrm{max}}, & \text{if $\Gamma_{k,n}$ is infeasible},\\
\Gamma_{k,n}, &\text{otherwise},
\end{array}\right.
\end{equation}
where $U_{\mathrm{max}}$ is a large constant indicating that the assignment of device $n$ and sub-channel $k$ is infeasible. In the one-to-one matching, the utility of any sub-channel is equal to the utility of the occupied device, i.e., $U_k(\Psi)=U_{\Psi(k)}(\Psi), \forall k\in\mathcal{K}$. By calculating the utility, the preference list of any player can be established. Since that problem \eqr{subproblem2} is a minimization problem, the preference of device $n$ can be defined below
\begin{equation}
(k,\Psi) \prec_n (k',\Psi') \Rightarrow U_n(\Psi)>U_n(\Psi').
\end{equation}
The above function indicates that device $n$ is willing to be assigned to sub-channel $k'$ in matching $\Psi'$, rather than $k$ in matching $\Psi$, since its utility can be strictly decreased by switching from $k$ to $k'$.

Since all players are matched, the considered case follows the concept of two-sided exchange matchings \cite{roth1992two}. In this case, if any device intends to be assigned to a sub-channel, it needs to exchange with the device occupying this sub-channel, instead of directly joining this sub-channel. A notation $\Psi_{n'}^n$ is introduced to represent the case where two devices $n$ and $n'$ are swapped in matching $\Psi$, which is defined as follows:
\begin{equation}
\Psi_{n'}^n(m)=\left\{
\begin{array}{ll}
\Psi(n')=k', & m=n,\\
\Psi(n)=k, & m=n',
\end{array}\right.
\end{equation}
where $\Psi_{n'}^n$ only replaces two pairs defined by $\Psi(n)=k$ and $\Psi(n')=k'$ into $\Psi(n)=k'$ and $\Psi(n')=k$, respectively. It is indicated that any swap operation involves four players, and hence, it should be approved by the involved players. However, in a matching with the incomplete preference list, there may be sub-channels that are unacceptable for some devices. In order to prioritize feasible combinations, the approval of sub-channels is removed, and the swap operation is approved by the swap-blocking pair $(n, n')$, which is defined below\cite{bodine2011peer}:
\begin{definition}\label{swapblocking}
A swap-blocking pair $(n, n')$ is confirmed if and only if the following conditions hold
\begin{enumerate}
\item $U_n(\Psi_{n'}^{n})\le U_n(\Psi)$ and $U_{n'}(\Psi_{n'}^{n})\le U_{n'}(\Psi)$;
\item At least one inequality above is strict. 
\end{enumerate}
\end{definition}

\begin{algorithm}[t]
\caption{Sub-Channel Assignment Algorithm}
\label{alg2}
\begin{algorithmic}[1]
\STATE \textbf{Initialization:}
\STATE Initialize initial matching $\Psi$ by randomly pairing all devices and sub-channels.
\STATE \textbf{Main Loop:}
\FOR{$n\in\mathcal{N}_t$}
\STATE Device $n$ makes a proposal to exchange with device $n'\in\mathcal{N}_t$, where $n\neq n'$.
\IF{$(n, n')$ is a swap-blocking pair}
\STATE Matching $\Psi_{n'}^n$ is approved.
\STATE Devices $n$ and $n'$ exchange sub-channels.
\STATE Set $\Psi=\Psi_{n'}^n$
\ENDIF
\ENDFOR
\STATE The main loop is repeated until no swap-blocking pair can be found in a complete round.
\end{algorithmic}
\end{algorithm}

By searching swap-blocking pairs, a matching based sub-channel assignment algorithm is presented in \textbf{Algorithm \ref{alg2}}. The proposed algorithm can be started from any initial matching. During the execution of the algorithm, an active device $n$ is selected and sequential swap is attempted with all other devices. If a swap-blocking pair $(n, n')$ is formed, the new matching $\Psi_{n'}^{n}$ is recorded, and the algorithm continues. The main loop of \textbf{Algorithm \ref{alg2}} executes repeatedly. When the last device has searched for all other devices, the first device becomes the active device again. The algorithm ends if no swap blocking pair can be found in a full round of the main loop. At this stage, a stable matching is established, which satisfies the following definition \cite{roth1992two, bodine2011peer}:
\begin{definition}\label{2es}
A matching $\Psi$ is two-sided exchange-stable (2ES) if and only if there is no further swap-blocking pair.
\end{definition}
Note that some devices may not be assigned to feasible sub-channels in the stable matching, i.e., their utilities are equal to $U_\mathrm{max}$. In this case, these devices cannot transmit local models to the server, and the corresponding sub-channel assignment indicators should be set to zero in the leader-level problem. In the following, the complexity, convergence and stability of the proposed matching based sub-channel assignment algorithm are analyzed. 

\subsubsection{Complexity}
By considering the worst case, the computational complexity of the proposed matching based algorithm can be expressed as $\mathcal{O}(CK^2)$, where $C$ is the number of main loops. In the worst case, $K$ devices can play the role of active devices to search $K-1$ devices, and thus $K(K-1)$ times of calculations should be performed in one loop.  With the given number of main loops $C$, the computational complexity of the proposed algorithm is obtained.

\subsubsection{Convergence}
Assuming $\Psi^{\text{(a)}}$ and $\Psi^{\text{(b)}}$ are two adjacent matching in \textbf{Algorithm \ref{alg2}}, i.e., $\Psi^{\text{(a)}}\to\Psi^{\text{(b)}}$, where $a\neq b$, one swap-blocking pair is found among this transformation. According to Definition \ref{swapblocking}, at least one device can achieve less utility, and the utilities of other devices cannot be increased. Hence, the matching transformation cannot be reversed. With the finite number of devices and sub-channels, the number of possible matchings is finite and equal to the Bell number \cite{ray2007game}. Therefore, from any initial matching, the proposed algorithm is guaranteed to converge to a stable matching.

\subsubsection{Stability}
Based on Definition \ref{2es}, any final matching obtained from the matching based sub-channel assignment algorithm is 2ES. Specifically, if the final matching obtained from Algorithm \ref{alg2} is not 2ES, there exists at least one swap-blocking pair, which can further reduce the sum utility of all devices. However, it contradicts the conditions for completing the proposed algorithm.
\section{Solution of Leader-Level Problem}\label{leadersolution}
To solve the formulated global loss minimization problem in \eqr{lproblem}, the loss function and local data must be available at the server, which is impractical and contradicts the motivation to utilize FL. In this section, by analyzing the impact of device selection on the expected convergence rate, such information can be detached from the leader-level problem. By employing the gradient descent method, with global model $\boldsymbol{w}^\mathrm{(t)}$, device $n$ can update the local model as follows:
\begin{equation}
\boldsymbol{w}_n^\mathrm{(t)}=\boldsymbol{w}^\mathrm{(t)}-\frac{\lambda}{\beta_n}\sum_{i=1}^{\beta_n}\nabla\ell(\boldsymbol{w}^\mathrm{(t)}; \boldsymbol{x}_{n,i}, y_{n,i}),
\end{equation}
where $\lambda$ is the learning rate. After that, the selected devices transmit the updated local models to the server for aggregation. By including device selection and sub-channel assignment, the aggregated global model in round $t$ is given by\footnote{Although we consider federated averaging (FedAvg) for aggregation in this paper, an extension of the results to other federated optimization algorithms \cite{wang2020tackling, li2020federated, hamer2020fedboost, wen2022fl} is straightforward.}
\begin{align}\nonumber
\boldsymbol{w}^\mathrm{(t+1)}\!=&\frac{\sum_{n=1}^{N}\!S_n^\mathrm{(t)}\!\sum_{k=1}^K\!\psi_{k,n}^\mathrm{(t)}\beta_n\boldsymbol{w}_{n}^\mathrm{(t)}}{\sum_{n=1}^{N}\!S_n^\mathrm{(t)}\!\sum_{k=1}^K\!\psi_{k,n}^\mathrm{(t)}\beta_n}\\\nonumber
=&\boldsymbol{w}^\mathrm{(t)}\!-\!\frac{\lambda\!\sum_{n=1}^{N}\!S_n^\mathrm{(t)}\!\sum_{k=1}^K\!\psi_{k,n}^\mathrm{(t)}\!\sum_{i=1}^{\beta_n}\!\!\nabla\ell(\boldsymbol{w}^\mathrm{(t)}; \boldsymbol{x}_{n,i}, y_{n,i})}{\sum_{n=1}^{N}\!S_n^\mathrm{(t)}\!\sum_{k=1}^K\!\psi_{k,n}^\mathrm{(t)}\beta_n}\\
=&\boldsymbol{w}^\mathrm{(t)}-\lambda[\nabla F(\boldsymbol{w}^\mathrm{(t)})-\hat{\boldsymbol{w}}^\mathrm{(t)}],
\label{globalmodelupdate}
\end{align}
where
\begin{equation}
\hat{\boldsymbol{w}}^\mathrm{(t)}\triangleq \nabla F(\boldsymbol{w}^\mathrm{(t)})\!-\!\frac{\sum_{n=1}^{N}\!S_n^\mathrm{(t)}\!\sum_{k=1}^K\!\psi_{k,n}^\mathrm{(t)}\!\sum_{i=1}^{\beta_n}\!\!\nabla\ell(\boldsymbol{w}^\mathrm{(t)}; \boldsymbol{x}_{n,i}, y_{n,i})}{\sum_{n=1}^{N}\!S_n^\mathrm{(t)}\!\sum_{k=1}^K\!\psi_{k,n}^\mathrm{(t)}\beta_n}.  
\end{equation}
In order to derive the expected convergence rate, the following assumptions are considered \cite{wang2019fl, samarakoon2020fl, chen2021fl1, mahdi2023cl}:
\begin{enumerate}[leftmargin=*]
\item With respect to the global model $\boldsymbol{w}^\mathrm{(t)}$, $\nabla F(\boldsymbol{w}^\mathrm{(t)})$ is uniformly Lipschitz continuous, i.e.,
\begin{equation}\label{assumption1}
\|\nabla F(\boldsymbol{w}^\mathrm{(t+1)})-\nabla F(\boldsymbol{w}^\mathrm{(t)})\|\le L\|\boldsymbol{w}^\mathrm{(t+1)}-\boldsymbol{w}^\mathrm{(t)}\|,
\end{equation}
where $L$ is a positive parameter.
\item $F(\boldsymbol{w}^\mathrm{(t)})$ is strongly convex with a positive constant $\mu$, i.e.,
\begin{equation}
F(\boldsymbol{w}^\mathrm{(t+1)})\ge F(\boldsymbol{w}^\mathrm{(t)})+(\boldsymbol{w}^\mathrm{(t+1)}-\boldsymbol{w}^\mathrm{(t)})^\top\nabla F(\boldsymbol{w}^\mathrm{(t)})+\frac{\mu}{2}\|\boldsymbol{w}^\mathrm{(t+1)}-\boldsymbol{w}^\mathrm{(t)}\|.
\label{assumption3}
\end{equation}
\item $F(\boldsymbol{w}^\mathrm{(t)})$ is twice-continuously differentiable, as below:
\begin{equation}\label{assumption4}
\mu\boldsymbol{I}\preceq\nabla^2F(\boldsymbol{w}^\mathrm{(t)})\preceq L\boldsymbol{I}.
\end{equation}
\item The following inequality is satisfied with any device $n$ and sample $i$:
\begin{equation}\label{assumption2}
\|\nabla \ell(\boldsymbol{w}^\mathrm{(t)});\boldsymbol{x}_{n,i},y_{n,i})\|^2\le\rho\|\nabla F(\boldsymbol{w}^\mathrm{(t)})\|^2,
\end{equation}
where $\rho$ is a non-negative constant.
\end{enumerate}
It is worth pointing out that the above assumptions are commonly considered in the FL related optimization, and can be satisfied by widely adopted loss functions. Based on these assumptions, the following proposition yields the effect of device selection on convergence rate.
\begin{proposition}\label{effect}
In the case that device selection and sub-channel assignment satisfy $S_n^\mathrm{(t)}=\sum_{k=1}^K\psi_{k,n}^\mathrm{(t)}, \forall n\in\mathcal{N}$, with the learning rate $\lambda=1/L$ and the optimal global model $\boldsymbol{w}^*$, the upper bound of convergence rate in round $t$ is
\begin{align}
\mathbb{E}[F(\boldsymbol{w}^\mathrm{(t+1)})\!-\!F(\boldsymbol{w}^*)]\!&\le\!\left(1\!-\!\frac{\mu}{L}\right)^t\!\mathbb{E}[F(\boldsymbol{w}^\mathrm{(1)})\!-\!F(\boldsymbol{w}^*)]\!\!\!\\\nonumber
&\quad+\!\frac{2\rho}{L}\sum_{i=1}^t\!\left(1\!-\!\frac{\mu}{L}\right)^{t-i}\!\frac{\|\nabla F(\boldsymbol{w}^\mathrm{(i)})\|^2}{\!\sum_{n=1}^{N}\!\beta_n}\!\!\sum_{n=1}^{N}\!\beta_n\!\!\left(\!1\!-\!S_n^\mathrm{(i)}\!\sum_{k=1}^K\!\psi_{k,n}^\mathrm{(i)}\!\right).
\end{align}
\begin{IEEEproof}
Refer to Appendix~C.
\end{IEEEproof}
\end{proposition}
It is indicated that the convergence rate is bounded by two terms. The first term, i.e., $\left(1\!-\!\frac{\mu}{L}\right)^t\mathbb{E}[F(\boldsymbol{w}^\mathrm{(1)})-F(\boldsymbol{w}^*)]$, is the expected gap between the first global loss and the optimal global loss in round $t$. The second term is related to device selection, where the status of devices in round $i$, i.e., $S_n^\mathrm{(i)}$, is included. Proposition \ref{effect} indicates the expected gap between the global loss in round $t$ and the optimal loss. In this case, the global loss minimization problem in \eqr{lproblem} can be achieved by minimizing the expected gap. Since that the first term does not include device selection, it can be treated as a constant. Meanwhile, terms $\frac{2\rho}{L}$ and $\frac{\|\nabla F(\boldsymbol{w}^\mathrm{(i)})\|^2}{\sum_{n=1}^{N}\beta_n}$ are positive and not effected by device selection, and hence, they can be removed. By incorporating the AoU based weight $\alpha_n^\mathrm{(t)}$ into the device selection indicator $S_n^\mathrm{(t)}$, the objective function of the leader-level problem \eqr{lproblem} can be reformulated as follows:
\begin{equation}
\min_{\mathbf{S}^\mathrm{(t)}} \!\quad\! \sum_{n=1}^{N}\beta_n\!\!\left(\!1\!-\!\alpha_n^\mathrm{(t)}S_n^\mathrm{(t)}\sum_{k=1}^K\psi_{k,n}^\mathrm{(t)}\!\right)
\end{equation}
By removing the constant term $\sum_{n=1}^{N}\beta_n$, the leader-level problem \eqr{lproblem} can be equivalently transformed as follows:
\begin{align}
\max_{\mathbf{S}^\mathrm{(t)}} \quad & \sum_{n=1}^{N}\alpha_n^\mathrm{(t)}\beta_n S_n^\mathrm{(t)}\sum_{k=1}^K\psi_{k,n}^\mathrm{(t)}\label{lproblem2}\\\nonumber
\textrm{s.t.} \quad & \text{(\ref{lproblem}a), (\ref{lproblem}b)}.
\end{align}

\begin{algorithm}[!t]
\caption{Device Selection Algorithm}
\label{alg3}
\begin{algorithmic}[1]
\STATE \textbf{Initialization:}
\STATE Generate list $\mathcal{Q}^\mathrm{(t)}$ based on \eqr{listq}.
\STATE Initialize set $\mathcal{N}_t$ by selecting the first $K$ devices from $\mathcal{Q}^\mathrm{(t)}$.
\STATE \textbf{Main Loop:}
\STATE Obtain sub-channel assignment from \textbf{Algorithm 2}.
\IF{$\sum_{n\in\mathcal{N}_t}\sum_{k=1}^K \psi_{k,n}^\mathrm{(t)}<K$ \textbf{and} $(N)\notin\mathcal{N}_t$}
\FOR{$n\in\mathcal{N}_t$}
\IF{$\sum_{k=1}^K \psi_{k,n}^\mathrm{(t)}=0$}
\STATE Remove device $n$ from set $\mathcal{N}_t$.
\STATE Add the next unselected device from list $\mathcal{Q}^\mathrm{(t)}$.
\ENDIF
\ENDFOR
\ENDIF
\end{algorithmic}
\end{algorithm}

\noindent Problem \eqr{lproblem2} can be treated as a weighted device selection problem, where AoU and data size play the role of weight factors. That is, the server tends to select the devices with large AoU and/or data size. In this case, the server can order and select devices based on this weight. In any round $t$, all devices are sorted in a list $\mathcal{Q}^\mathrm{(t)}$, which satisfies the following condition:
\begin{equation}\label{listq}
\alpha_{(1)}^\mathrm{(t)}\beta_{(1)} \ge \alpha_{(2)}^\mathrm{(t)}\beta_{(2)} \ge \dots \ge \alpha_{(N)}^\mathrm{(t)}\beta_{(N)},
\end{equation}
where $(1)$ and $(N)$ denote the devices with the highest and lowest priority, respectively. Moreover, it is indicated by problem \eqr{lproblem2} that the server tends to select more devices, and hence, constraint (\ref{lproblem}b) can be rewritten as $\sum_{n=1}^{N}S_n^\mathrm{(t)}=K$. According to the list $\mathcal{Q}^\mathrm{(t)}$, problem \eqr{lproblem2} can be solved by \textbf{Algorithm \ref{alg3}}. In the proposed device selection algorithm, $K$ devices with the highest priority are selected in the initialization phase. During \textbf{Algorithm \ref{alg3}}, any device not assigned to a sub-channel will be replaced, until all selected devices have been assigned to sub-channels, or all devices in list $\mathcal{Q}^\mathrm{(t)}$ have been adopted. In the worst case, all devices in list $\mathcal{Q}^\mathrm{(t)}$ are traversed, and therefore the complexity of \textbf{Algorithm \ref{alg3}} can be expressed as $\mathcal{O}(N)$.

\section{Simulation Results}
In this section, the performance of the proposed solution is simulated and demonstrated. In this simulation, multiple devices are randomly distributed in a disc with radius $R$, and the server is located in the center of the disc. The experiments includes three datasets (MNIST, CIFAR-10, and SST-2) with different models\footnote{For MNIST digit recognition tasks, a multi-layer perceptron neural network is built, where two ReLu hidden layers with $128$ and $256$ neurons follows by a softmax output layer. For CIFAR-10 image classification tasks, a convolutional neural network (CNN) is constructed with two 3x3 convolution layers (one with $32$ filters and one with $64$ filters, each followed by a 2x2 max pooling layer), a $128$-neuron ReLu hidden layer, and a softmax output layer. For SST-2 text classification tasks, a tokenizer with a vocabulary size of $4,000$ is included, and the neural network is built with a $128$-neuron ReLu hidden layer and a sigmoid output layer.}. The imbalanced independent identically distributed (IID) data distribution is adopted, where a factor $c_n\in[1,10]$ is randomly generated for all devices, and training samples ($500$ for MNIST, $50000$ for CIFAR-10, and $67349$ for SST-2) are shuffled and partitioned across devices based on fraction $c_n/\sum_{i\in\mathcal{N}} c_i$. Unlike \cite{wang2019fl, chen2021fl1, hamdi2022fl, liu2022fl, chen2021fl2, vu2020fl, chen2021flmatching, ji2021flmec} which focus on minimizing either global loss or latency, this paper studies the interaction between global loss minimization and latency minimization. Thereby, a direct comparison with the existing schemes in these works is unfair. Alternatively, the proposed scheme is benchmarked against the following conventional device selection schemes:
\begin{itemize}[leftmargin=*]
\item{\emph{AoU based DS}:} The server selects the top $K$ devices in~\eqr{listq}.
\item{\emph{Random DS}:} The server selects $K$ devices randomly.
\item{\emph{Cluster based DS}:} All devices are randomly allocated to $N/K$ clusters such that the clusters are selected in rotation.
\item{\emph{Fixed DS}:} The same $K$ devices are selected in all rounds.
\end{itemize}
In the above schemes, the proposed solutions are adopted, including monotonic optimization based resource allocation (MO-RA) and matching based sub-channel assignment (M-SA). Moreover, fixed resource allocation (FIX-RA) and random sub-channel assignment (R-SA) are incorporated as benchmarks, where $\tau_{k,n}=p_{k,n}=0.5, \forall k,n$ is set in FIX-RA. The parameters of the simulation are shown in Table \ref{parameter}.

\begin{table}[!t]
\centering
\caption{Table of Parameters}
\label{parameter}
\begin{tabular}{|l|c|}
\hline
Carrier frequency $f$ & $1$~GHz\\ \hline
AWGN noise power $\sigma^2$ & $-174$~dBm\\ \hline
Path loss exponent $a$ & $3.76$\\ \hline
Bandwidth for each sub-channel $B$ & $1$~MHz\\ \hline
Power consumption coefficient $\kappa_0$ & $10^{-28}$\\ \hline
CPU cycles for each bit of tasks $\mu$ & $10^7$\\ \hline
Computational capacity $C_n$ & $1$~GHz\\ \hline
Error tolerance $\epsilon $ & $0.01$\\ \hline
Model size $D(\boldsymbol{w})$ (MNIST/CIFAR-10/SST-2) & $1/5/5$~Mbit\\ \hline
Maximum energy $E_{n}^{\text{max}}$ (MNIST/CIFAR-10/SST-2)\!\!\! & $0.02/0.1/0.1$~joule\\ \hline
Learning rate $\lambda$ (MNIST/CIFAR-10/SST-2) & $0.01/0.001/0.01$ \\ \hline
Batch size (MNIST/CIFAR-10/SST-2) & $32/512/128$ \\ \hline
Optimizer (MNIST/CIFAR-10/SST-2) & SGD/Adam/SGD \\ \hline
\end{tabular}
\end{table}

\begin{figure*}[htp]
\centering{
\subfigure[MNIST]{\includegraphics[width=58mm]{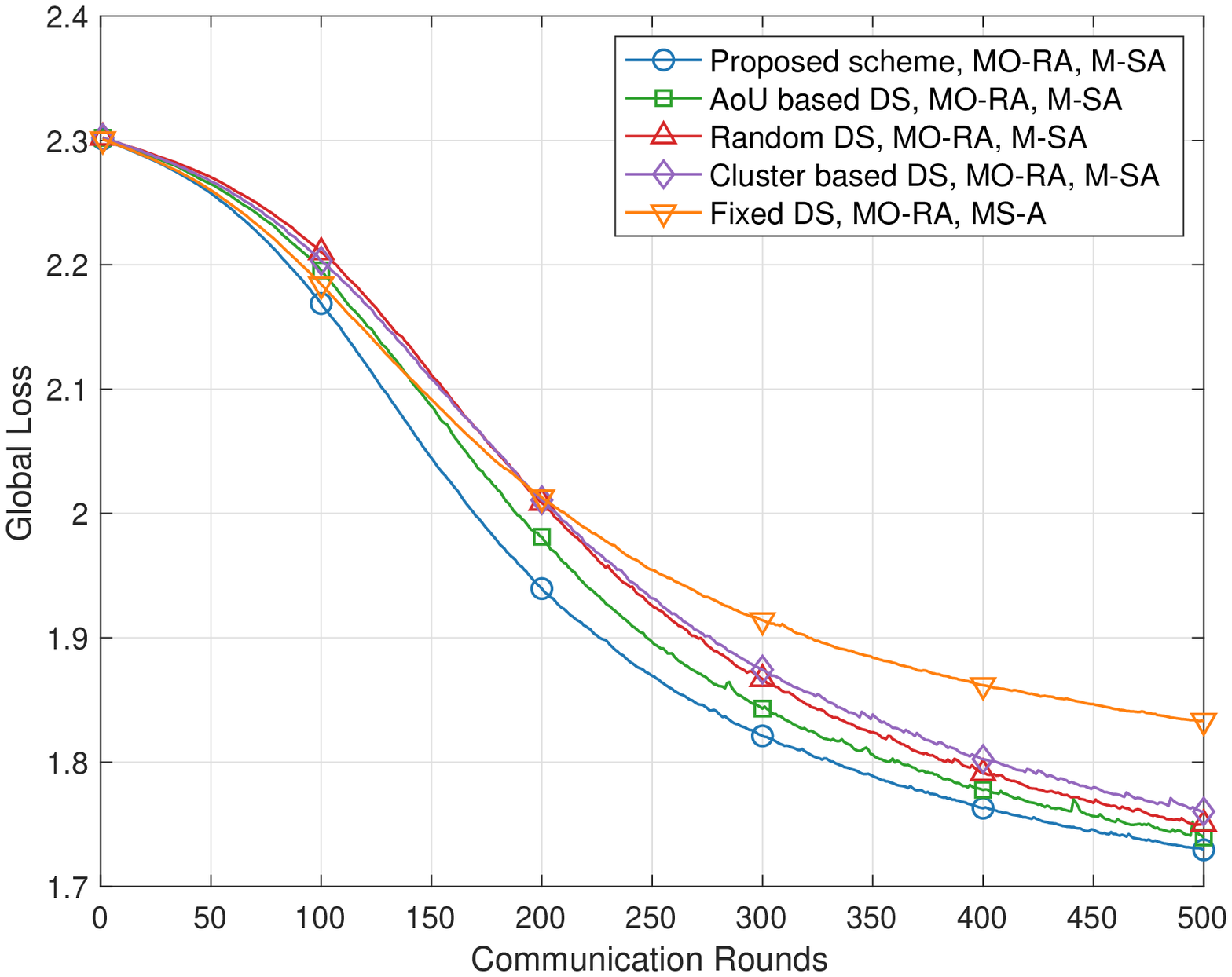}}\!\!\!\!\!\!\!\!\!
\subfigure[CIFAR-10]{\includegraphics[width=58mm]{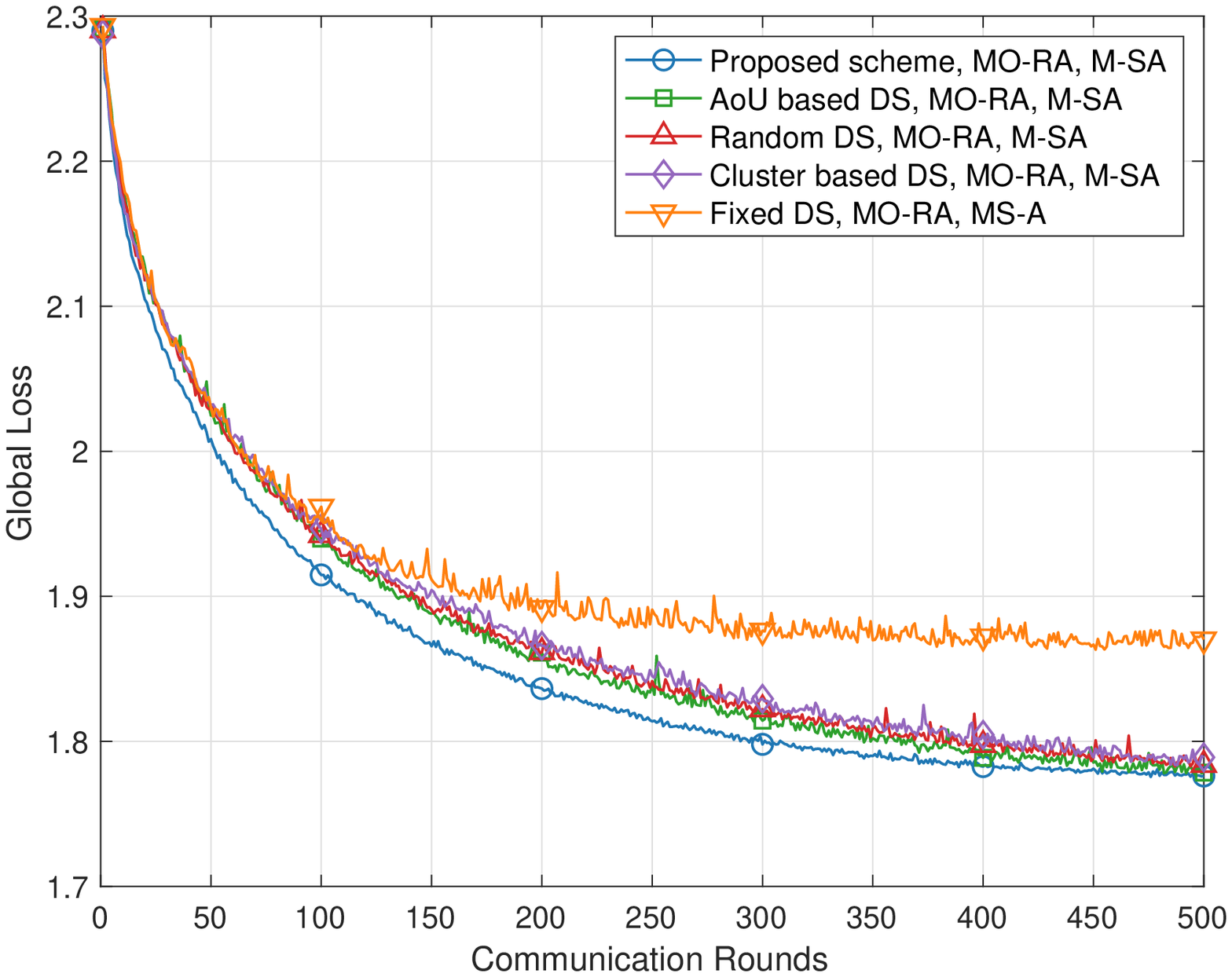}}\!\!\!\!\!\!\!\!\!
\subfigure[SST-2]{\includegraphics[width=58mm]{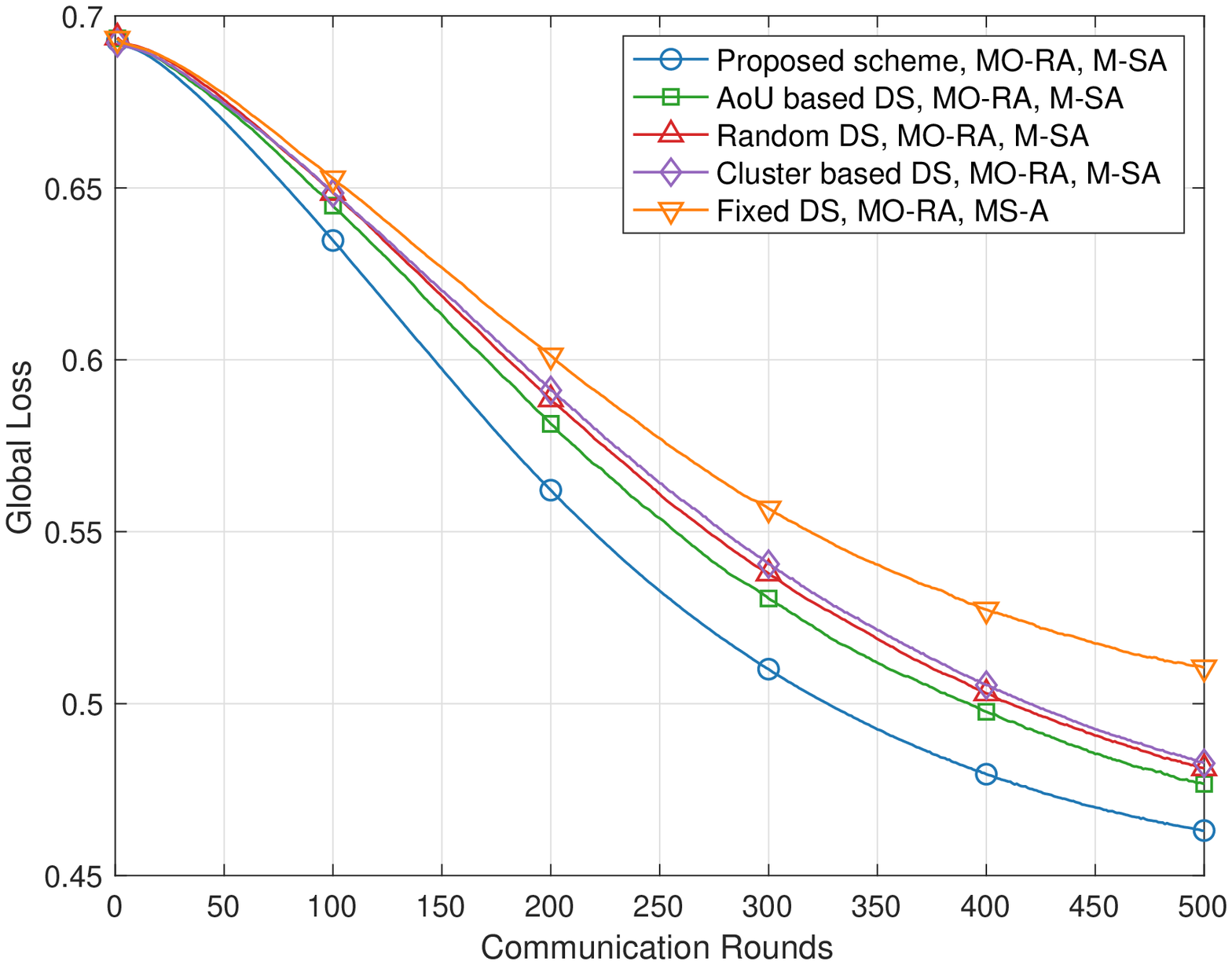}}}\vspace{-2mm}
\caption{The performance of the proposed scheme. $N=20$, $K=4$, $P_t=10$~dBm, and $R=500$~m.}
\label{result1}
\vspace{-4mm}
\end{figure*}

In \fref{result1}, the global loss achieved by different schemes is presented. It is indicated that with the limited number of sub-channels, the proposed scheme can achieve the lowest global loss on all datasets. This improvement benefits from two approaches, including AoU based weights and efficient utilization of sub-channels. The former is able to select devices that can provide more contributions in each communication round, while the latter ensures the maximization of the number of selected devices. Moreover, it can be observed that without  Algorithm \ref{alg3}, AoU based DS can still outperform other schemes, which confirms that AoU based weights have the capability to improve learning efficiency. In terms of the random DS and cluster based DS, the devices have the same probability to be selected, and hence, the similar global loss is achieved by these two schemes. For the fixed DS, due to the fact that the size of training data in this case is less than that in other schemes, its performance is the worst. Compared to the MNIST dataset, the CIFAR-10 image classification task is more complex, and hence, the differences between the schemes are not that obvious, as shown in \fref{result1}(b). However, the performance of these schemes is still consistent with \fref{result1}(a). It is worth noting that by employing the CIFAR-10 dataset, the data size and model size is increased, and therefore, the number of devices that can satisfy the energy constraint is reduced. This issue is severe when utilizing AoU based DS, as the server tends to select devices that are generally more difficult to meet this condition, resulting in poor performance. This effect is mitigated when a simpler task is adopted, as shown in \fref{result1}(c). Since there are only two labels on the SST-2 dataset, the differences between schemes become significant, and the advantages of the proposed scheme in global loss are clearly demonstrated.

\begin{figure*}[htp]
\centering{
\subfigure[MNIST]{\includegraphics[width=58mm]{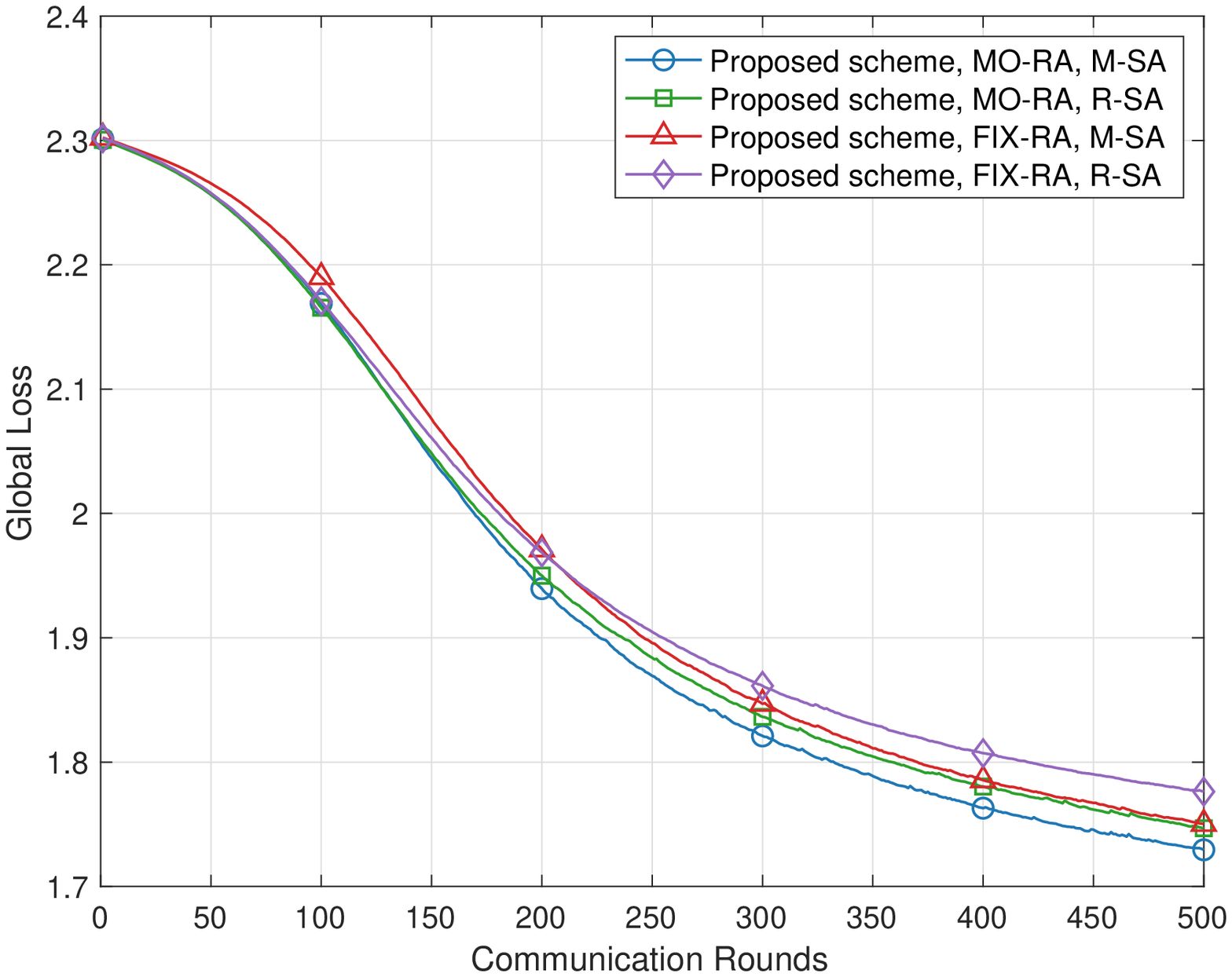}}\!\!\!\!\!\!\!\!\!
\subfigure[CIFAR-10]{\includegraphics[width=58mm]{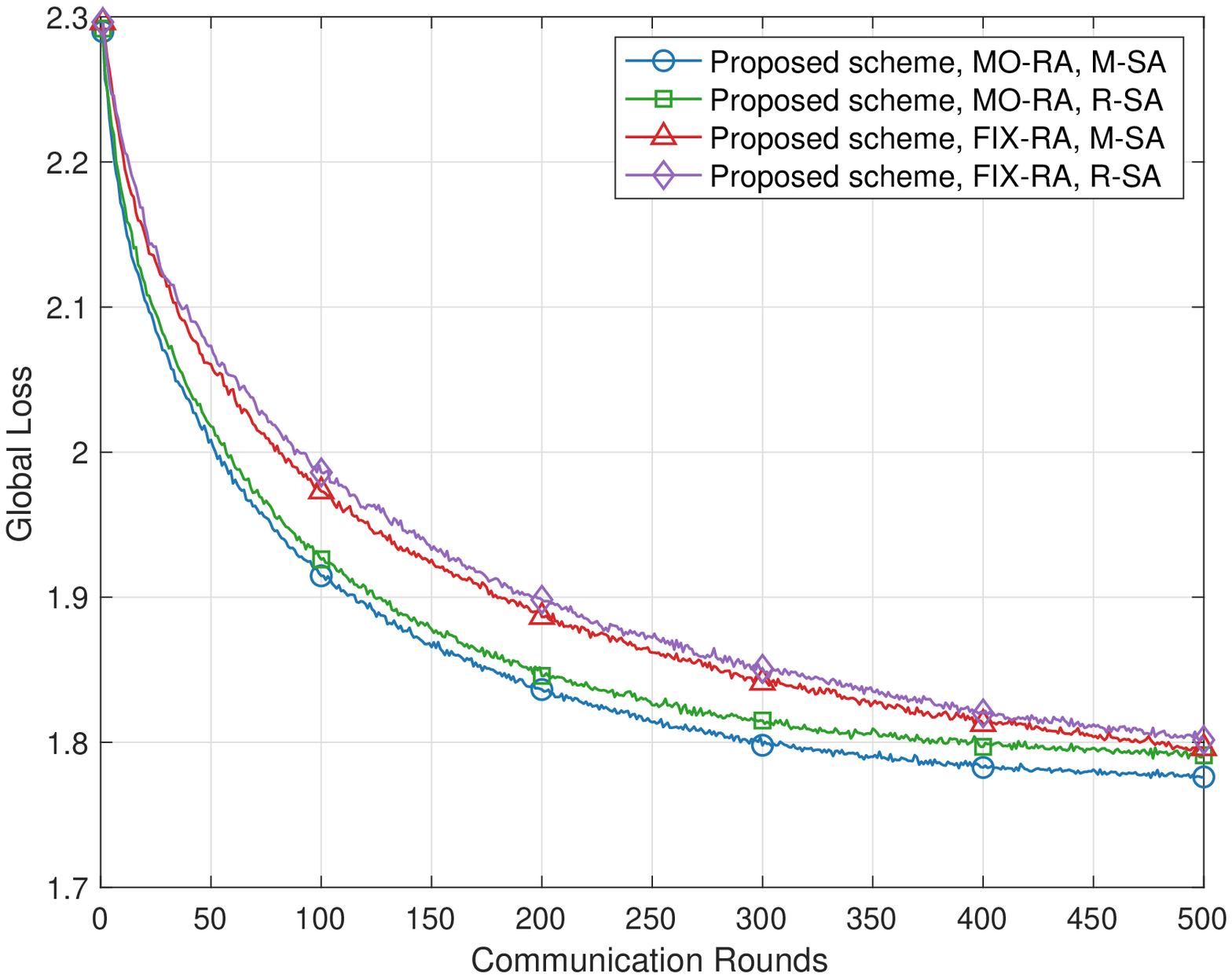}}\!\!\!\!\!\!\!\!\!
\subfigure[SST-2]{\includegraphics[width=58mm]{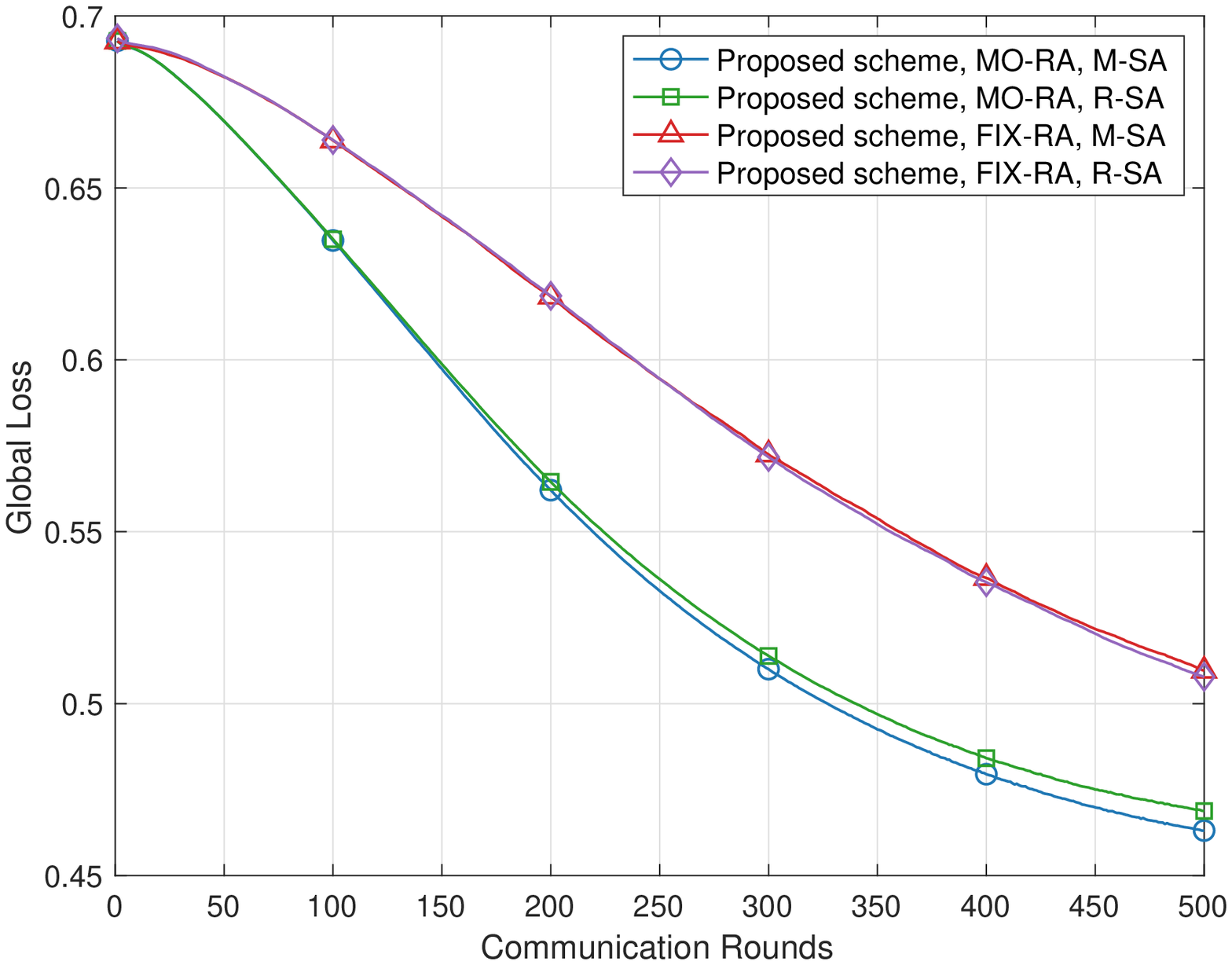}}}\vspace{-2mm}
\caption{The performance of the proposed scheme using different solutions. $N=20$, $K=4$, $P_t=10$~dBm, and $R=500$~m.}
\label{result2}
\vspace{-4mm}
\end{figure*}

\fref{result2} shows the performance of the proposed optimization solutions, where different resource allocation and sub-channel assignment approaches are incorporated into the proposed device selection scheme. It is clear that the proposed scheme can achieve the best performance with the proposed solutions, i.e., MO-RA and M-SA. This is because in this case, the server can select devices who can provide more contributions in the aggregation without replacing them with lower priority devices in \eqr{listq}. When other solutions are utilized, some devices may no longer meet the maximum energy consumption constraint and thus be replaced by devices with lower priority in \eqr{listq}, resulting in poor learning performance. This result also corroborates our conclusion that the contribution of a device in aggregation is proportional to its AoU and data size. In \fref{result2}(a), an obvious feature is that the proposed scheme using FIX-RA and R-SA performs well in the early stage of training (from round $1$ to round $100$), but poorly in the later stage. This is due to the fact that in this case, the number of devices that can satisfy the maximum energy consumption requirement is very small, and thus device selection is performed on a smaller subset. For simple tasks, such as MNIST with IID data distribution, repeatedly using smaller training datasets can achieve faster convergence in the early stages, however, this can lead to overfitting problems and poor final results. Moreover, it is worth pointing out that compared to sub-channel assignment, resource allocation plays a more important role in minimizing the global loss, and this trend is more significant when the model size is larger, as shown in \fref{result2}(b) and \fref{result2}(c).

\begin{figure}[t]
\centering{\includegraphics[width=84mm]{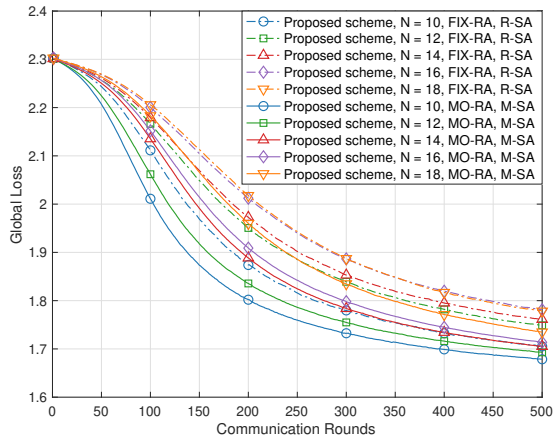}}
\caption{The impact of the number of devices with the MNIST dataset. $K=4$, $P_t=10$~dBm, and $R=500$~m.}
\label{resultN}
\vspace{-4mm}
\end{figure}

\begin{figure}[t]
\centering{\includegraphics[width=84mm]{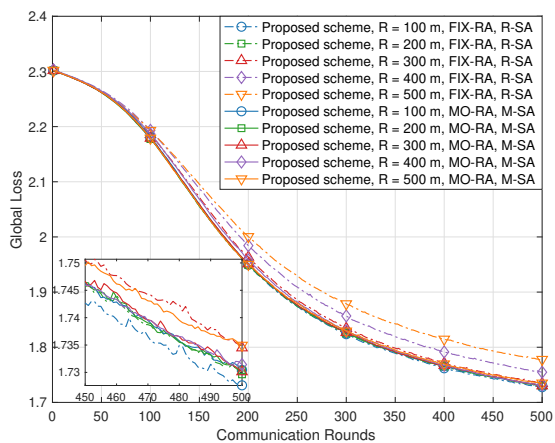}}
\caption{The impact of the radius with the MNIST dataset. $N=20$, $K=4$, and $P_t=10$~dBm.}
\label{resultradius}
\vspace{-4mm}
\end{figure}

In \fref{resultN} and \fref{resultradius}, the impact of the number of devices and radius on global loss is presented, respectively. In \fref{resultN}, since the size of the total training data is fixed, an increase in the number of devices means less training data per device. That is, when the number of selected devices, i.e., $K$, is fixed, as the number of devices rises, the amount of training data utilized in each communication round reduces, resulting in an increase in the global loss. With the proposed optimization solutions, devices that can provide more contributions can be selected, and therefore the learning performance is improved. In \fref{resultradius}, the increase in radius can be understood as a deterioration in channel conditions. According to Proposition \ref{infeasible}, in this case, more devices become unavailable and cannot participate in the aggregation. As a result, the achievable global loss increases with the radius. By employing the proposed optimization solutions, the negative impact caused by channel degradation can be alleviated to a certain extent, thereby narrowing the gap in global loss.

\begin{figure}[t]
\centering{\includegraphics[width=84mm]{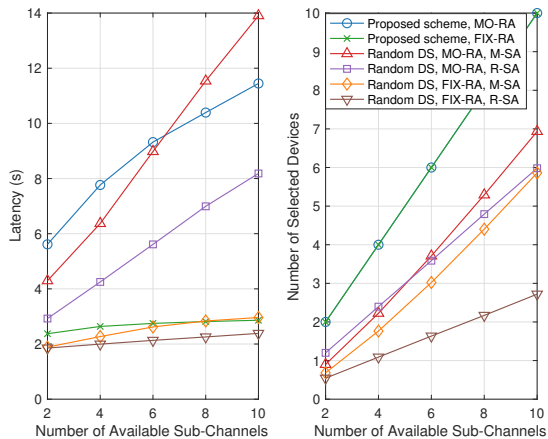}}
\caption{The impact of the number of available sub-channels. $P_t=10$~dBm, and $E_{n}^{\mathrm{max}}=0.02$~joule.}
\label{resultK}
\vspace{-4mm}
\end{figure}

In the considered Stackelberg game based framework, the number of selected devices varies between schemes, and therefore, the latency cannot be compared independently. It can be seen from \fref{resultK} that the proposed scheme can efficiently utilize all available sub-channels, while the latency of each communication round is also increased accordingly. That is because the proposed scheme needs to guarantee the performance of training through the leader level problem. This explains why the achievable global loss of the proposed scheme is lower than others. It is also indicated that the proposed RA and SA algorithms can increase the number of selected devices with random DS.

\begin{figure}[t]
\centering{\includegraphics[width=84mm]{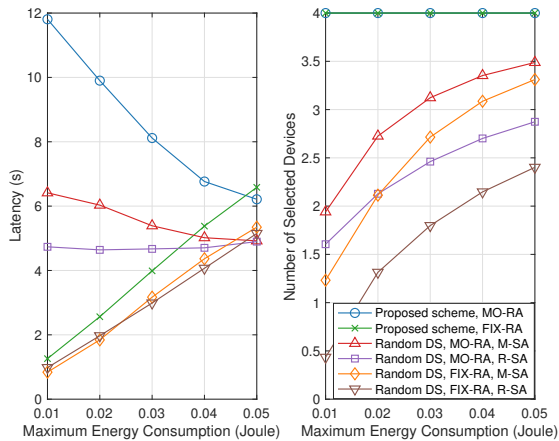}}
\caption{The impact of the maximum energy consumption. $K=4$, and $P_t=10$~dBm.}
\label{resultEmax}
\vspace{-4mm}
\end{figure}

The effect of energy limitation is demonstrated in \fref{resultEmax}, and the results confirm Proposition \ref{infeasible}. According to Proposition \ref{infeasible}, as the maximum energy consumption increases, the device participation increases. As a result, by employing random DS, the number of selected devices in each communication round is increased, and meanwhile the latency of each round also increases. On the other hand, it can be observed that the proposed algorithm, MO-RA, has the ability to dynamically adjust the computational resource allocation coefficients and power allocation coefficients. Therefore, the latency can be reduced with this solution.

\begin{figure}[t]
\centering{\includegraphics[width=84mm]{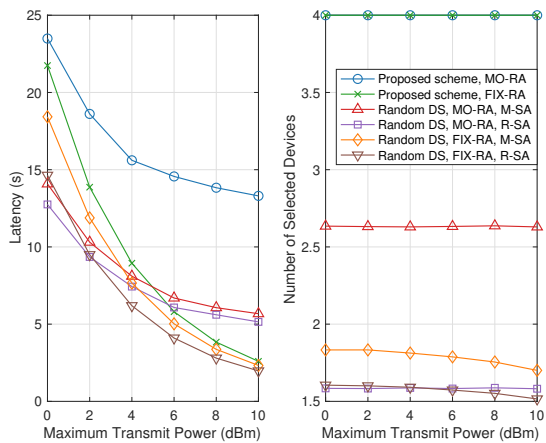}}
\caption{The impact of the maximum transmit power. $K=4$, and $E_{n}^{\mathrm{max}}=0.02$~joule.}
\label{resultPt}
\vspace{-4mm}
\end{figure}

As shown in \fref{resultPt}, the latency decreases with the increasing transmit power, since the achievable data rate can be increased. With FIX-RA, the number of selected devices is reduced when the transmit power is greater than $6$~dBm, because the fixed power allocation coefficient can no longer satisfy the energy consumption constraint. For MO-RA, the power allocation coefficient is optimized, and thus the number of selected devices is not affected. However, the decrease in latency has been slowed down accordingly.
\section{Conclusions}
This paper investigates FL in a practical wireless communication scenario, where limited sub-channels and energy consumption are considered. With the goal of minimizing convergence time, we jointly formulate global loss minimization and latency minimization based on the Stackelberg game. We present a monotonic optimization based algorithm to find the optimal solution of computational resource allocation and power allocation. Based on the obtained solution of resource allocation, we further develop a matching based algorithm with incomplete preference list to solve the sub-channel assignment problem. For the global loss minimization problem, we incorporate AoU based weights and derive an upper bound on convergence rate, which presents a priority for selecting devices. Simulation results demonstrate significant reduction of the global loss, establishing that our developed solutions can efficiently utilize available sub-channels and reduce latency. It is worth highlighting that the AoU based device selection can be extended to mobile environments as a practical and promising research direction.
\section*{Appendix~A: Proof of Proposition~\ref{infeasible}}
Suppose that device $n$ is selected and assigned to sub-channel $k$ in round $t$, i.e., $S_n^\mathrm{(t)}=1$ and $\psi_{k,n}^\mathrm{(t)}=1$, based on constraint (\ref{fproblem}a), the following condition will be satisfied by the optimal solutions $\tau_{k,n}^*$ and $p_{k,n}^*$:
\begin{equation}
 E_{k,n}^{\mathrm{cp}}(\tau_{k,n}^*)+E_{k,n}^{\mathrm{cm}}(p_{k,n}^*) \le E_{n}^{\mathrm{max}}.
\end{equation}
Due to the fact that the energy consumption for computation is strictly greater than zero, i.e., $ E_{k,n}^{\mathrm{cp}}(\tau_{k,n}^*)>0$, the following inequality can be obtained:
\begin{equation}
E_{k,n}^{\mathrm{cm}}(p_{k,n}^*)<E_{n}^{\mathrm{max}},
\end{equation}
which can be rewritten as follows:
\begin{equation}
\frac{p_{k,n}^*P_tD(\boldsymbol{w}_n^\mathrm{(t)})}{B\log_2(1\!+\!p_{k,n}^*|h_{k,n}|^2)}<E_{n}^{\mathrm{max}}.
\end{equation}
Since $p_{k,n}^*|h_{k,n}|^2>0$, $p_{k,n}^*|h_{k,n}|^2>\ln(1+p_{k,n}^*|h_{k,n}|^2)$ holds, the following inequality can be obtained:
\begin{equation}
\frac{\ln(2)P_tD(\boldsymbol{w}_n^\mathrm{(t)})}{B|h_{k,n}|^2}<\frac{p_{k,n}^*P_tD(\boldsymbol{w}_n^\mathrm{(t)})}{B\log_2(1+p_{k,n}^*|h_{k,n}|^2)}.
\end{equation}
Therefore, the following condition must be satisfied:
\begin{equation}
\frac{\ln(2)P_tD(\boldsymbol{w}_n^\mathrm{(t)})}{B|h_{k,n}|^2}<E_{n}^{\mathrm{max}},
\end{equation}
and this proposition is proved.\QEDA
\section*{Appendix~B: Proof of Proposition~\ref{monotonic}}
It is obvious that the time consumption of any device $n$ in sub-channel $k$ is monotonically decreasing with $\tau_{k,n}$ and $p_{k,n}$, and hence, the proof of this part is omitted. In terms of the energy consumption, as shown in constraint (\ref{subproblem11}a), the monotonicity can be proved by two parts. The first term, $E_{k,n}^{\mathrm{cp}}(\tau_{k,n})$, is the energy consumption for computation, which is monotonically increasing with the computational resource allocation coefficient. The second term is the energy consumption for communication, which can be presented by
\begin{equation}\label{energymono}
E_{k,n}^{\mathrm{cm}}(p_{k,n})=\frac{p_{k,n}P_tD(\boldsymbol{w}_n^\mathrm{(t)})}{B\log_2(1+p_{k,n}|h_{k,n}|^2)}.
\end{equation}
The derivative of the above function is
\begin{align}
\frac{\partial E_{k,n}^{\mathrm{cm}}(p_{k,n})}{\partial p_{k,n}}\!&=\!\frac{\ln(2)P_tD(\boldsymbol{w}_n^\mathrm{(t)})}{B(1\!+\!p_{k,n}|h_{k,n}|^2)\!\ln^2(1\!+\!p_{k,n}|h_{k,n}|^2)}\!\!\!\\\nonumber
&\quad\times\![(1\!+\!p_{k,n}|h_{k,n}|^2)\!\ln(1\!+\!p_{k,n}|h_{k,n}|^2)\!-\!p_{k,n}|h_{k,n}|^2].
\end{align}
Since the first term of the above function is always greater than zero, the monotonicity of function $E_{k,n}^{\mathrm{cm}}(p_{k,n})$ depends on the term in the square bracket. Particularly, $E_{k,n}^{\mathrm{cm}}(p_{k,n})$ is an increasing function if the following inequality holds:
\begin{equation}
(1\!+\!p_{k,n}|h_{k,n}|^2)\!\ln(1\!+\!p_{k,n}|h_{k,n}|^2)\!-\!p_{k,n}|h_{k,n}|^2\ge 0.
\end{equation}
Suppose $\theta_{k,n}\triangleq \frac{1}{p_{k,n}|h_{k,n}|^2+1}$, the above inequality is equivalent to the following inequality:
\begin{equation}
\frac{1}{\theta_{k,n}}(\theta_{k,n}-1-\ln \theta_{k,n})\ge 0.
\end{equation}
Since $\theta_{k,n}>0$, $\ln \theta_{k,n}\le \theta_{k,n}-1$ holds, and then the above inequality is always satisfied, which indicates that \eqr{energymono} is an increasing function. As a result, the total energy consumption is an increasing function, and this proposition is proved.\QEDA
\section*{Appendix~C: Proof of Proposition~\ref{effect}}
In order to prove this proposition, an auxiliary function is defined as follows:
\begin{equation}\label{auxiliaryfunction}
G(\boldsymbol{w})=\frac{L}{2}\boldsymbol{w}^\top\boldsymbol{w}-F(\boldsymbol{w}).
\end{equation}
With respect to $\boldsymbol{w}$, the second-order partial derivative of the above function is given by
\begin{equation}
\frac{\partial^2 G(\boldsymbol{w})}{\partial\boldsymbol{w}^2}=L-\frac{\partial^2 F(\boldsymbol{w})}{\boldsymbol{w}^2}.
\end{equation}
Since $F(\boldsymbol{w})$ satisfies the uniformly Lipschitz condition, the above function is always greater than or equal to zero, and hence, $G(\boldsymbol{w})$ is a convex function. By utilizing the second-order Taylor series expansion, the following inequality can be obtained:
\begin{equation}
G(\boldsymbol{w}^\mathrm{(t+1)})\ge G(\boldsymbol{w}^\mathrm{(t)})+(\boldsymbol{w}^\mathrm{(t+1)}-\boldsymbol{w}^\mathrm{(t)})^\top\nabla G(\boldsymbol{w}^\mathrm{(t)}).
\end{equation}
From \eqr{auxiliaryfunction}, the above inequality can be equivalently transformed as follows:
\begin{equation}
F(\boldsymbol{w}^\mathrm{(t+1)})\le F(\boldsymbol{w}^\mathrm{(t)})\!+\!(\boldsymbol{w}^\mathrm{(t+1)}\!-\!\boldsymbol{w}^\mathrm{(t)})^\top\nabla F(\boldsymbol{w}^\mathrm{(t)})\!+\!\frac{L}{2}\|\boldsymbol{w}^\mathrm{(t+1)}\!-\!\boldsymbol{w}^\mathrm{(t)}\|^2.
\end{equation}
Based on \eqr{globalmodelupdate}, the following inequality can be obtained:
\begin{equation}
F(\boldsymbol{w}^\mathrm{(t+1)}) \le F(\boldsymbol{w}^\mathrm{(t)})\!-\!\lambda[\nabla F(\boldsymbol{w}^\mathrm{(t)})\!-\!\hat{\boldsymbol{w}}^\mathrm{(t)}]^\top \nabla F(\boldsymbol{w}^\mathrm{(t)})\!+\!\frac{\lambda^2L}{2}\|\nabla F(\boldsymbol{w}^\mathrm{(t)})\!-\!\hat{\boldsymbol{w}}^\mathrm{(t)}\|^2.
\end{equation}
When $\lambda=1/L$, the above inequality can be transformed as follows:
\begin{align}\nonumber
\mathbb{E}[F(\boldsymbol{w}^\mathrm{(t+1)})] &\le \mathbb{E}\left\{F(\boldsymbol{w}^\mathrm{(t)})\!-\!\lambda[\nabla F(\boldsymbol{w}^\mathrm{(t)})\!-\!\hat{\boldsymbol{w}}^\mathrm{(t)}]^\top\nabla F(\boldsymbol{w}^\mathrm{(t)})\!+\!\frac{\lambda^2L}{2}\|\nabla F(\boldsymbol{w}^\mathrm{(t)})\!-\!\hat{\boldsymbol{w}}^\mathrm{(t)}\|^2\right\}\\\nonumber
&=\mathbb{E}\left[F(\boldsymbol{w}^\mathrm{(t)})\!-\!\frac{1}{L}\|\nabla F(\boldsymbol{w}^\mathrm{(t)})\|^2\!+\!\frac{1}{L}(\hat{\boldsymbol{w}}^\mathrm{(t)})^\top\nabla F(\boldsymbol{w}^\mathrm{(t)})\right.\\\nonumber
&\quad \left.+\frac{1}{2L}\|\nabla F(\boldsymbol{w}^\mathrm{(t)})\|^2\!-\!\frac{1}{L}(\hat{\boldsymbol{w}}^\mathrm{(t)})^\top\nabla F(\boldsymbol{w}^\mathrm{(t)})\!+\!\frac{1}{2L}\|\hat{\boldsymbol{w}}^\mathrm{(t)}\|^2 \right]\\
&= \mathbb{E}[F(\boldsymbol{w}^\mathrm{(t)})]\!-\!\frac{1}{2L}\|\nabla F(\boldsymbol{w}^\mathrm{(t)})\|^2\!+\!\frac{1}{2L}\mathbb{E}(\|\hat{\boldsymbol{w}}^\mathrm{(t)}\|^2).
\label{expectfg}
\end{align}
By defining
\begin{align}
\left\{\begin{array}{ll}
f(\mathcal{N}_t,\boldsymbol{\psi}^\mathrm{(t)},\boldsymbol{w}^\mathrm{(t)})&\!\triangleq\!\sum\limits_{n\in\mathcal{N}_t}\!\sum\limits_{k=1}^K\!\psi_{k,n}^\mathrm{(t)}\!\sum\limits_{i=1}^{\beta_n}\!\nabla\ell(\boldsymbol{w}^\mathrm{(t)}; \boldsymbol{x}_{n,i}, y_{n,i}),\vspace{1mm}\\
g(\mathcal{N}_t,\boldsymbol{\psi}^\mathrm{(t)},\boldsymbol{w}^\mathrm{(t)})&\!\triangleq\!\sum\limits_{n\in\mathcal{N}_t}\!\sum\limits_{k=1}^K\!\psi_{k,n}^\mathrm{(t)}\!\sum\limits_{i=1}^{\beta_n}\!\|\nabla\ell(\boldsymbol{w}^\mathrm{(t)}; \boldsymbol{x}_{n,i}, y_{n,i})\|,
\end{array}\right.
\end{align}
based on condition $S_n^\mathrm{(t)}=\sum_{k=1}^K\psi_{k,n}^\mathrm{(t)}=1, \forall n\in\mathcal{N}_t$, the following equation can be derived:
\begin{align}\nonumber
\mathbb{E}(\|\hat{\boldsymbol{w}}^\mathrm{(t)}\|^2)\!&=\!\mathbb{E}\!\!\left[\bigg\| \nabla F(\boldsymbol{w}^\mathrm{(t)})\!-\!\frac{f(\mathcal{N}_t,\boldsymbol{\psi}^\mathrm{(t)},\boldsymbol{w}^\mathrm{(t)})}{\sum_{n=1}^{N}\!S_n^\mathrm{(t)}\!\sum_{k=1}^K\!\psi_{k,n}^\mathrm{(t)}\beta_n}\bigg\|^2\right]\\
&=\!\mathbb{E}\!\!\left[\bigg\|\frac{f(\mathcal{N}_t,\boldsymbol{\psi}^\mathrm{(t)},\boldsymbol{w}^\mathrm{(t)})}{\sum_{n=1}^{N}\!\beta_n}\!\!-\!\!\frac{f(\mathcal{N}_t,\boldsymbol{\psi}^\mathrm{(t)},\boldsymbol{w}^\mathrm{(t)})}{\sum_{n=1}^{N}\!S_n^\mathrm{(t)}\!\sum_{k=1}^K\!\psi_{k,n}^\mathrm{(t)}\beta_n}\!\!+\!\!\frac{\sum_{n\in\mathcal{N}\backslash\mathcal{N}_t}\!\sum_{i=1}^{\beta_n}\!\!\nabla\ell(\boldsymbol{w}^\mathrm{(t)}; \boldsymbol{x}_{n,i}, y_{n,i})}{\sum_{n=1}^{N}\beta_n}\bigg\|^2\right],\!\!\!
\end{align}
where $\mathcal{N}\backslash\mathcal{N}_t$ is the collection of unselected devices in round $t$. According to the triangle-inequality, the above equation can be transformed as follows:
\begin{align}\nonumber
\mathbb{E}(\|\hat{\boldsymbol{w}}^\mathrm{(t)}\|^2)&\le\mathbb{E}\!\left\{\frac{\left[\sum_{n=1}^{N}\!\beta_n\!\!\left(\!1\!-\!S_n^\mathrm{(t)}\!\sum_{k=1}^K\!\psi_{k,n}^\mathrm{(t)}\!\right)\right]\!g(\mathcal{N}_t,\boldsymbol{\psi}^\mathrm{(t)},\boldsymbol{w}^\mathrm{(t)})}{(\sum_{n=1}^{N}\!\beta_n)(\sum_{n=1}^{N}\!S_n^\mathrm{(t)}\!\sum_{k=1}^K\!\psi_{k,n}^\mathrm{(t)}\beta_n)}\right.\\
&\quad\left.+\frac{\sum_{n\in\mathcal{N}\backslash\mathcal{N}_t}\!\sum_{i=1}^{\beta_n}\!\|\nabla\ell(\boldsymbol{w}^\mathrm{(t)}; \boldsymbol{x}_{n,i}, y_{n,i})\|}{\sum_{n=1}^{N}\beta_n}\right\}^2.\!\!\!
\label{expectg1}
\end{align}
According to \eqr{assumption2}, inequalities
\begin{equation}
g(\mathcal{N}_t,\boldsymbol{\psi}^\mathrm{(t)},\boldsymbol{w}^\mathrm{(t)})\!\le\!\sum_{n=1}^{N}\!S_n^\mathrm{(t)}\!\sum_{k=1}^K\!\psi_{k,n}^\mathrm{(t)}\beta_n\!\sqrt{\rho\|\nabla F(\boldsymbol{w}^\mathrm{(t)})\|^2},
\end{equation}
and
\begin{equation}
\sum_{n\in\mathcal{N}\backslash\mathcal{N}_t}\!\sum_{i=1}^{\beta_n}\|\nabla\ell(\boldsymbol{w}^\mathrm{(t)}; \boldsymbol{x}_{n,i}, y_{n,i})\|\le \left[\sum_{n=1}^{N}\!\beta_n\!\!\left(\!1\!-\!S_n^\mathrm{(t)}\!\sum_{k=1}^K\!\psi_{k,n}^\mathrm{(t)}\!\right)\!\right]\!\!\sqrt{\rho\|\nabla F(\boldsymbol{w}^\mathrm{(t)})\|^2},
\end{equation}
can be obtained. Therefore, \eqr{expectg1} can be rewritten as follows:
\begin{align}\nonumber
\mathbb{E}(\|\hat{\boldsymbol{w}}^\mathrm{(t)}\|^2)&\!\le\!\mathbb{E}\!\left\{\!\!\frac{\left[\!\sum_{n=1}^{N}\!\beta_n\!\!\left(\!1\!\!-\!\!S_n^\mathrm{(t)}\!\sum_{k=1}^K\!\psi_{k,n}^\mathrm{(t)}\!\right)\!\right]\!\!\sqrt{\rho\|\nabla F(\boldsymbol{w}^\mathrm{(t)})\|^2}}{\sum_{n=1}^{N}\!\beta_n}\right.\\\nonumber
&\!\!\quad\left.+\frac{\left[\!\sum_{n=1}^{N}\!\beta_n\!\!\left(\!1\!-\!S_n^\mathrm{(t)}\!\sum_{k=1}^K\!\psi_{k,n}^\mathrm{(t)}\!\right)\!\right]\!\!\sqrt{\rho\|\nabla F(\boldsymbol{w}^\mathrm{(t)})\|^2}}{\sum_{n=1}^{N}\!\beta_n}\!\right\}^2\\
&=\!\frac{4\rho\|\nabla F(\boldsymbol{w}^\mathrm{(t)})\|^2}{\left(\!{\sum_{n=1}^{N}\!\beta_n}\!\right)^2}\mathbb{E}\!\!\left[\sum_{n=1}^{N}\!\beta_n\!\!\left(\!1\!-\!S_n^\mathrm{(t)}\!\sum_{k=1}^K\!\psi_{k,n}^\mathrm{(t)}\!\right)\!\right]^2.
\label{expectg2}
\end{align}
Due to the fact that the number of all devices' samples is not less than the number of unselected devices' samples, i.e.,
\begin{equation}
\sum_{n=1}^{N}\beta_n\ge\sum_{n=1}^{N}\beta_n\!\!\left(\!1\!-\!S_n^\mathrm{(t)}\!\sum_{k=1}^K\!\psi_{k,n}^\mathrm{(t)}\!\right)\ge 0,
\end{equation}
the following inequality can be obtained from \eqr{expectg2}:
\begin{equation}
\mathbb{E}(\|\hat{\boldsymbol{w}}^\mathrm{(t)}\|^2)\!\le\!\frac{4\rho\|\nabla F(\boldsymbol{w}^\mathrm{(t)})\|^2}{\sum_{n=1}^{N}\!\beta_n}\mathbb{E}\!\left[\sum_{n=1}^{N}\!\beta_n\!\!\left(\!1\!-\!S_n^\mathrm{(t)}\!\sum_{k=1}^K\!\psi_{k,n}^\mathrm{(t)}\!\right)\!\right].
\end{equation}
As a result, \eqr{expectfg} can be rewritten as follows:
\begin{align}
\mathbb{E}[F(\boldsymbol{w}^\mathrm{(t+1)})]\le\mathbb{E}[F(\boldsymbol{w}^\mathrm{(t)})]-\frac{1}{2L}\|\nabla F(\boldsymbol{w}^\mathrm{(t)})\|^2\!+\!\frac{2\rho\|\nabla F(\boldsymbol{w}^\mathrm{(t)})\|^2}{L\!\sum_{n=1}^{N}\!\beta_n}\!\sum_{n=1}^{N}\beta_n\!\left(\!1\!-\!S_n^\mathrm{(t)}\!\sum_{k=1}^K\!\psi_{k,n}^\mathrm{(t)}\!\right).
\end{align}
By subtracting $\mathbb{E}[F(\boldsymbol{w}^*)]$ in both sides of the above function, the following inequality can be obtained
\begin{align}\nonumber
\mathbb{E}[F(\boldsymbol{w}^\mathrm{(t+1)})\!-\!F(\boldsymbol{w}^*)]\!&\le\!\mathbb{E}[F(\boldsymbol{w}^\mathrm{(t)})\!-\!F(\boldsymbol{w}^*)]\!-\!\frac{1}{2L}\|\nabla F(\boldsymbol{w}^\mathrm{(t)})\|^2\\
&\quad+\!\frac{2\rho\|\nabla F(\boldsymbol{w}^\mathrm{(t)})\|^2}{L\!\sum_{n=1}^{N}\!\beta_n}\!\sum_{n=1}^{N}\beta_n\!\left(\!1\!-\!S_n^\mathrm{(t)}\!\sum_{k=1}^K\!\psi_{k,n}^\mathrm{(t)}\!\right).\!\!
\label{oneround}
\end{align}
According to \cite{boyd2004convex}, the following inequality can be obtained from \eqr{assumption3} and \eqr{assumption4}:
\begin{equation}
\|\nabla F(\boldsymbol{w}^\mathrm{(t)})\|^2\ge 2\mu[F(\boldsymbol{w}^\mathrm{(t)})-F(\boldsymbol{w}^{*})].
\end{equation}
Hence, \eqr{oneround} can be rewritten as follows:
\begin{align}\nonumber
\mathbb{E}[F(\boldsymbol{w}^\mathrm{(t+1)})\!-\!F(\boldsymbol{w}^*)]\!&\le\!\left(1\!-\!\frac{\mu}{L}\right)\mathbb{E}[F(\boldsymbol{w}^\mathrm{(t)})\!-\!F(\boldsymbol{w}^*)]\\
&\quad+\!\frac{2\rho\|\nabla F(\boldsymbol{w}^\mathrm{(t)})\|^2}{L\!\sum_{n=1}^{N}\!\beta_n}\!\!\sum_{n=1}^{N}\beta_n\!\left(\!1\!-\!S_n^\mathrm{(t)}\!\sum_{k=1}^K\!\psi_{k,n}^\mathrm{(t)}\!\right).\!\!
\end{align}
Similarly, the convergence of successive rounds is given by
\begin{align}\nonumber
\mathbb{E}[F(\boldsymbol{w}^\mathrm{(t+1)})\!-\!F(\boldsymbol{w}^*)]\!&\le\!\left(1\!-\!\frac{\mu}{L}\right)^2\mathbb{E}[F(\boldsymbol{w}^\mathrm{(t-1)})\!-\!F(\boldsymbol{w}^*)]\\\nonumber
&\quad+\!\left(1\!-\!\frac{\mu}{L}\right)\!\frac{2\rho\|\nabla F(\boldsymbol{w}^\mathrm{(t-1)})\|^2}{L\!\sum_{n=1}^{N}\!\beta_n}\!\!\sum_{n=1}^{N}\beta_n\!\left(\!1\!-\!S_n^\mathrm{(t-1)}\!\sum_{k=1}^K\!\psi_{k,n}^\mathrm{(t-1)}\!\right)\\
&\quad+\!\frac{2\rho\|\nabla F(\boldsymbol{w}^\mathrm{(t)})\|^2}{L\!\sum_{n=1}^{N}\!\beta_n}\!\!\sum_{n=1}^{N}\beta_n\!\left(\!1\!-\!S_n^\mathrm{(t)}\!\sum_{k=1}^K\!\psi_{k,n}^\mathrm{(t)}\!\right).
\end{align}
As a result, we derive the upper bound of the convergence rate as
\begin{align}
\mathbb{E}[F(\boldsymbol{w}^\mathrm{(t+1)})\!-\!F(\boldsymbol{w}^*)]\!&\le\!\left(1\!-\!\frac{\mu}{L}\right)^t\!\mathbb{E}[F(\boldsymbol{w}^\mathrm{(1)})\!-\!F(\boldsymbol{w}^*)]\\\nonumber
&\quad+\!\!\frac{2\rho}{L}\sum_{i=1}^t\!\left(1\!-\!\frac{\mu}{L}\right)^{t-i}\!\frac{\|\nabla F(\boldsymbol{w}^\mathrm{(i)})\|^2}{\!\sum_{n=1}^{N}\!\beta_n}\!\!\sum_{n=1}^{N}\beta_n\!\left(\!1\!-\!S_n^\mathrm{(i)}\!\sum_{k=1}^K\!\psi_{k,n}^\mathrm{(i)}\!\right),
\end{align}
which completes the proof.\QEDA
\bibliographystyle{IEEEtran}
\bibliography{KaidisBib}
\end{document}